\documentclass{article} % For LaTeX2e
\usepackage{iclr2026_conference,times}

\usepackage{amsmath,amsfonts,bm}

\def\eqref#1{equation~\ref{#1}}
\def\1{\bm{1}}

\DeclareMathAlphabet{\mathsfit}{\encodingdefault}{\sfdefault}{m}{sl}
\SetMathAlphabet{\mathsfit}{bold}{\encodingdefault}{\sfdefault}{bx}{n}

\usepackage{hyperref}
\usepackage{url}

\usepackage{amsmath,amssymb,amsfonts}

\usepackage{wrapfig}
\usepackage{graphicx}
\usepackage{subcaption}
\usepackage{caption}
\usepackage{booktabs} % for professional looking tables
\usepackage{array}    % for better column formatting

\usepackage{amsthm} 
\newtheorem{theorem}{Theorem}

\newtheorem{proposition}[theorem]{Proposition}

\usepackage{booktabs}
\usepackage{adjustbox}

\usepackage{minted}
\usepackage{listings}
\usepackage{algorithm}
\usepackage[noend]{algpseudocode}
\algrenewcommand\alglinenumber[1]{\footnotesize #1:}

\usepackage{caption}
\usepackage{float}

\title{LLM Routing with Dueling Feedback}
\iclrfinalcopy % Uncomment for camera-ready version, but NOT for submission.

\author{Chao-Kai Chiang\textsuperscript{1} \quad
    Takashi Ishida\textsuperscript{2,1} \quad
    Masashi Sugiyama\textsuperscript{2,1} \quad
\\
    \textsuperscript{1}The University of Tokyo, Tokyo, Japan \\ 
    \textsuperscript{2}RIKEN AIP, Tokyo, Japan
\\
    \texttt{\{chaokai,ishi,sugi\}@k.u-tokyo.ac.jp} 
}

\newcommand{\new}{}

\newcommand{\mycmt}[1]{}

\newcommand{\eqary}[1]{\begin{align*}       % Equation array
#1
\end{align*}}
\newcommand{\num}{\refstepcounter{equation}\tag{\theequation}}
\begin{document}

\maketitle

\begin{abstract}
% \rcmt{The abstract must be limited to one paragraph.}
We study LLM routing, the problem of selecting the best model for each query while balancing user satisfaction, model expertise, and inference cost. We formulate routing as contextual dueling bandits, learning from pairwise preference feedback rather than absolute scores, thereby yielding label-efficient and dynamic adaptation. Building on this formulation, we introduce Category-Calibrated Fine-Tuning (CCFT), a representation-learning method that derives model embeddings from offline data using contrastive fine-tuning with categorical weighting. These embeddings enable the practical instantiation of Feel-Good Thompson Sampling for Contextual Dueling Bandits (FGTS.CDB), a theoretically grounded posterior-sampling algorithm. We propose four variants of the categorical weighting that explicitly integrate model quality and cost, and we empirically evaluate the proposed methods on the RouterBench and MixInstruct datasets. Across both benchmarks, our methods achieve lower cumulative regret and faster convergence, with better robustness and performance-cost balance than strong baselines built with a general-purpose OpenAI embedding model.
% \mycmt{Terminology: We propose a generic strategy called CCFT. Four variants/methods (\texttt{Perf}, \texttt{Perf\_cost}, \texttt{Excel\_perf\_cost}, \texttt{Excel\_mask}) are used interchangeably in the main text.}
\end{abstract}

\vspace{-1mm}\section{Introduction}
\label{sec:intro}
The potential of large language models (LLMs) is so great that they have become a necessary part of daily life, with applications ranging from office assistance and fashion/dining suggestions to entertainment.
% The potential of large language models (LLMs) is so large that it almost becomes a necessary part of daily life, with applications from office assistance, fashion/dining suggestions, to entertainment.
LLM routing refers to a problem of dynamically selecting the most suitable LLM from a set of candidates for each query in a sequence of questions.
Before the emergence of a universally dominant and affordable foundation model, routing is important because the choice of LLM must align with user traits, model expertise, and cost.
To balance these three key factors, cascading algorithms such as FrugalGPT \citep{chen2024frugalgpt} and AutoMix \citep{AggarwalMAPMZGR24} were first proposed.
The idea is to query a cheaper model first and advance the query to a more expensive one if the current response is unlikely to meet the user’s expectation.

A drawback of cascading is the accumulated cost and latency caused by calling multiple LLM candidates to generate the final response.
% A drawback of cascading is the accumulated cost and latency caused by a series of calls to multiple LLM candidates to obtain the final response to output.
To avoid this, supervised routing methods \citep{shnitzer2023llmrouting, lu2024zooter, ding2024hybridllm, hu2024routerbench, srivatsa2024harnessing} were proposed.
In general, a supervised router reduces the latency by applying a classification or a regression prediction before the LLM query.
The supervised approach has evolved into several branches. 
A branch studied ensembles \citep{jiang2023mixinst_blender, maurya-etal-2025-selectllm, zhang2025avengers, zhang2025avengerspro}, which allows the agent to select a subset or fuse answers. 
Another branch focused on cost-aware model assignment \citep{sakota2024forc, hu2024routerbench, liu2024optllm} to balance cost and performance.
There are also works that combine representation learning \citep{feng2025graphrouter, zhuang2025embedllm} to strengthen the user and model’s semantic information before training.

For supervised routing methods, having abundant real-valued annotations with high-quality label information is critical for successful classification or regression training.
Unfortunately, such a requirement is often unrealistic in the context of LLM routing.
In some cases, users are either reluctant or unmotivated to provide feedback. 
They may also be unable to quantify their satisfaction, especially when dealing with open-ended questions or when they lack the ability to verify the correctness of the LLM's response.
To ease the annotation burden, some efforts focused on weak supervision \citep{sugiyama2022weak, chiang2025unified}, which only collects binary feedback such as like/dislike or pairwise comparison \citep{ong2025routellm, zhao2024eagle, wang2025mixllm}.
The advantage is that one-click feedback is user-friendly, and it is more confident to say response A is better than B than to assign response A a score out of ten.
The reason that makes a weakly supervised approach appealing is that, as shown by the referred papers, the binary feedback can be translated into a model ranking or an estimate of the labeling function. 

In addition to the challenge of annotation, adaptivity remains a key challenge to be addressed in developing a usable routing system.
Shifts in query distributions, such as changes in trending topics like fashion or temporal variations between work hours and leisure time, introduce non-stationary conditions. 
Moreover, new LLMs and benchmarks are continuously introduced, resulting in a constantly evolving environment for routing.
Because of their static nature, supervised learning-based routing policies struggle to address multiple adaptation challenges simultaneously.
% \mycmt{(original) These evolving conditions often exceed the capacity of static, supervised learning-based routing policies.} 
Addressing such challenges in a dynamic environment is a key motivation for adopting online learning approaches for LLM routing, as these offer the ability to continuously learn and optimize the routing policy in real time.
The online algorithms adopted in prior work can be categorized into three classes: multi-armed bandits \citep{nguyen2024metallm, dai2024C2MAB-V, li2025llmbandit}, contextual bandits \citep{wang2025mixllm}, and reinforcement learning \citep{sikeridis2025pickllm}.

To build a practical routing system that fits the various requests, multiple challenges should be addressed simultaneously. 
However, we notice that little effort has been made to jointly solve the challenges of adaptivity and weak supervision, even if the community has already made significant progress in respective directions.
% However, we notice that there was little effort at jointly solving the challenges of adaptivity and weak supervision, even if the community has already made huge progress in respective directions.
To the best of our knowledge, \cite{wang2025mixllm} is the only attempt to address adaptivity and pointwise feedback (e.g., like/dislike) at the same time.
Therefore, this paper focuses on investigating LLM routing under a stochastic bandit setting, which captures the dynamics of a changing environment, and operates under weak supervision in the form of pairwise preference feedback (e.g., response A is preferred over response B).
% \mycmt{Replace question by query.}
The advantages of the project are threefold:
First, we introduce the Feel-Good Thompson Sampling for Contextual Dueling Bandits (FGTS.CDB) algorithm  \citep{li2024fgtscdb} as the core module, which naturally integrates weak supervision (dueling feedback) and adaptive learning (bandit algorithm) in both input and learning design, expanding methodological options for future research.
Second, FGTS.CDB is theoretically grounded, providing a clear explanation of how binary feedback relates to a utility function shaped by user satisfaction, model expertise, and cost.
Third, it offers a platform to analyze its strengths and limitations, enabling development of a practical LLM router that does not rely on high-quality annotations and remains robust in dynamic environments.
% The advantages of the project are threefold:
% First, we introduce the contextual dueling bandit algorithm FGTS.CDB \citep{li2024fgtscdb} as the core module.
% It naturally combines weak supervision (i.e., the dueling feedback) and adaptive learning (i.e., the bandit algorithm) in the algorithm input and the design of the learning procedure.
% Its introduction provides the community with broader methodological choices for future research.
% Second, it is a theoretically justified method that offers a clear explanation of how binary feedback is connected to the utility function that defines the model's performance, which is a complex function of user satisfaction, model expertise, and model cost.
% Third, it provides a platform for us to investigate what makes it work and what makes it fail, so that we can realize it to be a practical LLM router that does not require high-quality annotations and is robust to dynamically changing environments.

The paper’s contributions are summarized as follows.
% \mycmt{Make it THREE.}
\begin{itemize}
    \item We propose Category-Calibrated Fine-Tuning (CCFT), a generic embedding strategy to encode LLM expertise. The feature function from CCFT enables the first trainable contextual dueling learner for LLM routing.
    % We propose Category-Calibrated Fine-Tuning (CCFT), a generic embedding strategy to encode an LLM’s expertise.
    % The feature function constructed by CCFT allows us to build the first trainable contextual dueling learner in LLM routing.
    \item Strong evidence for the efficacy of the proposed strategy is provided by experiments on two real-world datasets, RouterBench \citep{hu2024routerbench} and MixInstruct \citep{jiang2023mixinst_blender}.
    Four CCFT variants are implemented and evaluated, and the cumulative regret curves show convergence to the optimal strategy, selecting the best-matching model for each query.
    % Four implementations of CCFT are tested and compared. 
    % The cumulative regret curves demonstrate that the proposed methods converge to the optimal strategy, which selects the best-matching model for each query.
    \item The proposed methods demonstrate robust generalization on the unseen benchmark and achieve a balance between performance and cost. 
    They incorporate common practices, including prompting, embedding model fine-tuning, and the use of both open-source and black-box text embedding models. 
    Therefore, the experiments contribute to the accumulation of substantial knowledge and expertise in addressing LLM routing challenges.
    % \item The proposed methods also demonstrate the merits of robust generalization to an unseen benchmark and performance-cost balance.
    % \item In the experiments, several common practices of LLM, including prompting, context fine-tuning, open-sourced and black-box text embedding models, are integrated with the proposed methods, accumulating substantial knowledge and skills of addressing LLM routing.
\end{itemize}

\vspace{-1mm}\section{Related Work}

\paragraph{LLM selection strategies}
% \label{sec:related_routing}
LLM selection can be organized along two axes: how candidates are queried and what learning signal is used.
On the querying side, cascading systems \citep{chen2024frugalgpt, AggarwalMAPMZGR24,narasimhan2024faster,chuang2024learning} issue a sequence of calls, starting from a cheap model and escalating to stronger ones until a confidence or quality threshold is met.
In contrast, one-shot routers predict a single or two target model(s) before inference.
One-shot routing is preferable when latency must remain small or when a diverse pool of models with complementary domain strengths is available \citep{jiang2023mixinst_blender} and we wish to select one  (or two, for preference feedback).
Within one-shot routing, offline methods train a classifier or regressor on a fixed labeled set to map queries to models \citep[e.g.,][]{shnitzer2023llmrouting, lu2024zooter, ding2024hybridllm, hu2024routerbench, srivatsa2024harnessing,jitkrittum2025universal}.
Online methods instead adapt the routing policy on the fly using bandits or reinforcement learning to cope with distribution shift and evolving model pools \citep[e.g.,][]{nguyen2024metallm, dai2024C2MAB-V, li2025llmbandit, wang2025mixllm, sikeridis2025pickllm}.

On the signal side, many routers rely on pointwise supervision, i.e., correct/incorrect or scalar ratings, while others leverage preference (pairwise) feedback that compares two candidates, which can be easier to elicit \citep{ong2025routellm, zhao2024eagle, wang2025mixllm}.
Our work lies in the online, one-shot setting with preference signals:~we model routing as contextual dueling bandits and instantiate a Thompson-sampling-style learner that updates from pairwise comparisons while balancing performance and cost.

% \bcmt{TI: Similar but different directions we are not currently discussing are model merging, model collaboration, and routing within mixture of experts}

% \paragraph{Preference Feedback and Performance-Cost Balance}
% \mycmt{(Maybe to Appendix when we have time? $\Rightarrow$ This is addressed in the paragraph above.)}

\paragraph{Contextual Dueling Bandits and Feel-Good Thompson Sampling}
% \label{sec:bandits}

The contextual bandit problem extends the classical multi-armed bandit setting by leveraging side information \citep{langford2007epochgreedy}.
It has found widespread applications in areas such as online advertising, recommender systems, and mobile health \citep{li2010contextualBanditNews, agarwal2016contextualdecisions, tewari2017contextualHealth}.
A widely used and empirically effective class of algorithms for contextual bandits is Thompson Sampling (TS) \citep{thompson1933TS}, known for its strong empirical performance \cite{chapelle2011empiricalTS, osband2017posterior}.
% Research on contextual dueling bandits has advanced along several directions. 
Research on contextual dueling bandits has taken several algorithmic and theoretical directions.
\cite{kumagai2017continuousdueling} analyzed dueling bandits with a continuous action space and, under strong convexity and smoothness, established dimension-free regret guarantees. 
Building on preference models, \cite{bengs2022stochasticcontextual} introduced the CoLSTIM algorithm for stochastic contextual dueling bandits under linear stochastic transitivity, providing learning guarantees tailored to this structure. 
Recently, \cite{di2024varianceaware} proposed VACDB, an action-elimination-based method that achieves tighter, variance-dependent regret bounds for contextual settings.

Feel-Good Thompson Sampling (FGTS) was proposed to reconcile TS's strong empirical performance with frequentist-style guarantees \citep{zhang2022fgts}. \cite{fan2023powerfgts} offered a unified analysis framework showing how FGTS yields robust guarantees across several linear contextual bandit variants. The FGTS idea has also been extended to reinforcement learning, e.g., Model-based Optimistic Posterior Sampling (MOPS) for Markov decision processes \citep{agarwal2022modelbasedops}. To the best of our knowledge, our work is the first to apply FGTS to LLM routing, connecting preference-based bandit principles with practical model-selection pipelines.

\vspace{-1mm}\section{Background}
\label{sec:background}

% \vspace{-2mm}\subsection{Feel-Good Thompson Sampling for Contextual Dueling Bandits}
% \mycmt{Need a section level symbol ($a_k$ vs $\rho_k$) refinement (next section, actually). $\Rightarrow$ See the 1st paragraph of \S~\ref{sec:failure_cases}.}
The contextual dueling bandit problem can be seen as a repeated game between a bandit algorithm and an environment for $T$ rounds. 
In each round $t = 1, 2, \ldots T$, the algorithm observes a contextual vector $x_t$ from the environment.
% In each round $t = 1, 2, \ldots T$, the algorithm observe a contextual vector $x_t$ representing the query from the environment.
Then, the algorithm selects two actions $a^1_t, a^2_t \in \mathcal{A} = \{a_k\}_{k=1}^{K}$ in response to the environment.
After presenting the responses, the algorithm observes a preference feedback $y_t$.
The performance of the algorithm is measured by its cumulative regret 
\eqary{
    \operatorname{Regret}(T) 
    := \sum_{t=1}^{T} \left[ r^{*}(x_t, a_t^{*}) - \frac{r^{*}(x_t, a_t^{1}) + r^{*}(x_t, a_t^{2})}{2} \right],
    \num\label{eq:regret}
}
where $r^{*}(x, a)$ is the utility function and $a_t^{*} = \arg\max_{a \in \mathcal{A}} r^{*}(x_t, a)$ is the best action for input $x_t$.

The setting fits seamlessly to the LLM routing problem studied in this paper if we view $x_t$ as the query embedding,
% that encodes query semantics and user traits, 
$a^1_t$ and $a^2_t$ as two LLMs, $y_t$ as the preference feedback, and the $r^{*}(x, a)$ as the function balancing the user satisfaction and model score\footnote{The model score is computed based on LLM performance, cost, latency, and other relevant factors.} that we want to optimize.
Minimizing $\operatorname{Regret}(T)$ precisely captures the goal of routing: identifying the optimal LLM $a_t^{*}$ at each round, rather than committing to a single fixed $a^{*}$ across all rounds.
The connection between the weak supervision of preference feedback $y$ and the ideal supervision provided by $r^{*}(x, a)$ is captured by the Bradley--Terry--Luce (BTL) model~\citep{hunter2004btl, luce2005btl}:
Given query $x$ and two LLMs $a^1$ and $a^2$, the probability of observing $a^1$ is preferred over $a^2$ (i.e., $y=1$) is
$$\mathbb{P}(y = 1 \mid x, a^1, a^2) 
    = 
    % \frac{\exp(r^{*}(x, a^1))}{\exp(r^{*}(x, a^1)) + \exp(r^{*}(x, a^2))}
    % = 
    \exp\big(-\sigma(r^{*}(x, a^1) - r^{*}(x, a^2))\big),$$
where $\sigma(z) = \log(1 + \exp(-z))$\footnote{
A more general setting accepting more than two candidates is called the Plackett-Luce (PL) model \citep{pmlr-v32-soufiani14, pmlr-v48-khetan16, ren2018pacranking}.
}.

% %%%%%%%%%%%%%%%%%%%%%%%%%%%%%%%%%%%%%%%%%%%%%%%%%%%%%%%%%%%%
% \begin{algorithm}
% \caption{FGTS.CDB}\label{alg:fgts-cdb}
% \begin{algorithmic}[1]
% \State Given action space $\mathcal{A}$ and hyperparameters $\eta, \mu$. Initialize $S_0 = \varnothing$.
% \For{$t = 1, \ldots, T$}
%   \State Receive query $x_t$.
%   \For{$j = 1, 2$}
%     \State Sample model parameter $\theta_t^{j}$ from the posterior distribution: 
%     \eqary{
%     \textstyle  % tip-TI
%         p^j(\theta \mid S_{t-1}) 
%         \propto 
%         \exp\left( -\sum_{i=1}^{t-1} L^j(\theta, x_i, a_i^1, a_i^2, y_i) \right) p_0(\theta)
%     }
%     % $p^{j}(\,\cdot \mid S_{t-1})$, defined in (4.1).
%     \State Select LLM $a_t^{j} = \arg\max_{a \in \mathcal{A}}\; \langle \theta_t^{j}, \,\phi(x_t,a) \rangle$ to generate the response.
%   \EndFor
%   \State Receive preference feedback $y_t$.
%   \State Update history $S_t \leftarrow S_{t-1} \cup \{(x_t, a_t^{1}, a_t^{2}, y_t)\}$.
% \EndFor
% \end{algorithmic}
% \end{algorithm}

Assuming the linear reward model $r^{*}(x, a) = \langle \theta^{*}, \phi(x, a) \rangle$, \cite{li2024fgtscdb} proposed Algorithm~\ref{alg:fgts-cdb} and proved that it achieves $\mathbb{E}[\operatorname{Regret}(T)] = \widetilde{\mathcal{O}}(d\sqrt{T})$, where $d$ is the dimension of $\theta$. 
FGTS.CDB learns the LLM selection strategies, $\theta^{1}_t$ and $\theta^{2}_t$, in an online fashion.
Its success hinges on the posterior $p^j(\theta \mid S_{t-1})$ defined by the likelihood function
\eqary{
    L^j(\theta, x, a^1, a^2, y) 
    = 
    \eta \, \sigma\big(y \langle \theta, \phi(x, a^1) - \phi(x, a^2) \rangle \big) - \mu \max_{\tilde{a} \in \mathcal{A}} \langle \theta, \phi(x, \tilde{a}) - \phi(x, a^{3-j}) \rangle.
    \num\label{eq:likelihood}
}
%%%%%%%%%%%%%%%%%%%%%%%%%%%%%%%%%%%%%%%%%%%%%%%%%%%%%%%%%%%%
\begin{wrapfigure}{r}{0.6\textwidth}
\vspace{-4mm}
\begin{minipage}{\linewidth}
    \begin{algorithm}[H]  % [H] = non-floating
    \caption{FGTS.CDB}\label{alg:fgts-cdb}
    \begin{algorithmic}[1]
\State Given action space $\mathcal{A}$ and hyperparameters $\eta, \mu$. Initialize $S_0 = \varnothing$.
\For{$t = 1, \ldots, T$}
    \State Receive query $x_t$.
    \For{$j = 1, 2$}
        \State Sample model parameter $\theta_t^{j}$ from posterior 
        \eqary{
        \textstyle  % tip-TI
            p^j(\theta \mid S_{t-1}) 
            \propto 
            \exp\left( -\sum_{i=1}^{t-1} L^j(\theta, x_i, a_i^1, a_i^2, y_i) \right) p_0(\theta)
        }   % $p^{j}(\,\cdot \mid S_{t-1})$, defined in (4.1).
        \State Select LLM $a_t^{j} = \arg\max_{a \in \mathcal{A}}\; \langle \theta_t^{j}, \,\phi(x_t,a) \rangle$ % to generate the response.
        \Statex \quad\quad\quad to generate the response.
    \EndFor
    \State Receive preference feedback $y_t$.
    \State Update history $S_t \leftarrow S_{t-1} \cup \{(x_t, a_t^{1}, a_t^{2}, y_t)\}$.
\EndFor
    \end{algorithmic}
    \end{algorithm}
\end{minipage}
\vspace{-2mm}
\end{wrapfigure}
The intuition behind $L^{j}(\cdot)$ is as follows: if $\theta$ aligns well with the preference feedback $y$, that is, when $y \langle \theta, \phi(x, a^1) - \phi(x, a^2) \rangle$ is positive and large, then $\theta$ is more likely to be chosen.
The second term in the likelihood function serves as a ``feel-good'' component, encouraging the selection of a $\theta$ that outperforms past selections made by the other selection strategy\footnote{Specifically, note that, $a^{3-j} = a^2$ for $j=1$ and $a^{3-j} = a^1$ for $j=2$.}. 
As time $t$ progresses, the observation history $S_t$ accumulates, allowing FGTS to dynamically adjust its samples $\theta^{1}_{t+1}$ and $\theta^{2}_{t+1}$ accordingly.
In an evolving environment, any changes are reflected in $S_t$, and adaptivity is inherently ensured by the bandit algorithm.

Being theoretically oriented, \cite{li2024fgtscdb} assumes that the feature function $\phi(x, a)$ is given and perfect. 
However, this assumption generally does not hold in real-world applications, as shown in the following section.

% \mycmt{\textbf{(Fine-tuning background)} This section talk about FGTS.CDB mainly. Add an appendix section for contrastive fine-tuning if the discussion in main text is not enough.}

% \vspace{-2mm}\subsection{Context Contrastive Fine-Tuning}
% \rcmt{Here or Appendix?}

% \vspace{-2mm}\subsection{Materials}

% \noindent\textbf{Stochastic Preference Model.} 
% In this work, we assume that the preference $y_t$ follows a Bernoulli distribution 
% according to the Bradley--Terry--Luce (BTL) model~\cite{Hunter2004,Luce2005}: 
% Given context $x_t$ and responses $a_t^1, a_t^2$, 
% the probability of $a_t^1$ being preferred over $a_t^2$ is
% \eqary{
%     \mathbb{P}(y_t = 1 \mid x_t, a_t^1, a_t^2) 
%     = 
%     \frac{\exp(r^{*}(x_t, a_t^1))}{\exp(r^{*}(x_t, a_t^1)) + \exp(r^{*}(x_t, a_t^2))}
%     = 
%     \exp\big(-\sigma(r^{*}(x_t, a_t^1) - r^{*}(x_t, a_t^2))\big),
% }
% where $\sigma(z) = \log(1 + \exp(-z))$\footnote{
% A more general setting accepting more than two candidates is called the Plackett-Luce (PL) model \cite{Soufiani et al., 2014; Khetan & Oh, 2016; Ren et al., 2018}.
% }.

% \noindent\textbf{Learning Objective.} 
% The goal is to minimize the cumulative average regret:
% \eqary{
%     \operatorname{Regret}(T) 
%     := \sum_{t=1}^{T} \left[ r^{*}(x_t, a_t^{*}) - \frac{r^{*}(x_t, a_t^{1}) + r^{*}(x_t, a_t^{2})}{2} \right],
% }
% where $a_t^{*} = \arg\max_{a \in \mathcal{A}} r^{*}(x_t, a)$ is the optimal LLM for question $x_t$.

\vspace{-1mm}\section{A Generic Representation Learning Strategy}
\label{sec:generic_method}
In \S~\ref{sec:failure_cases}, we demonstrate how an imperfect $\phi(x, a)$ can hinder learning and highlight a key obstacle in applying FGTS.CDB to the routing problem.
Subsequently, in \S~\ref{sec:proposed_model_emb}, we introduce our methods for constructing feature functions that enable FGTS.CDB to function effectively in practical LLM routing scenarios.

\vspace{-2mm}\subsection{The Failure of Naive Implementations}
\label{sec:failure_cases}
\begin{wrapfigure}{r}{0.4\textwidth}
\vspace{-8mm}
\begin{minipage}{0.4\textwidth}
    \begin{center}
    \includegraphics[width=\linewidth]{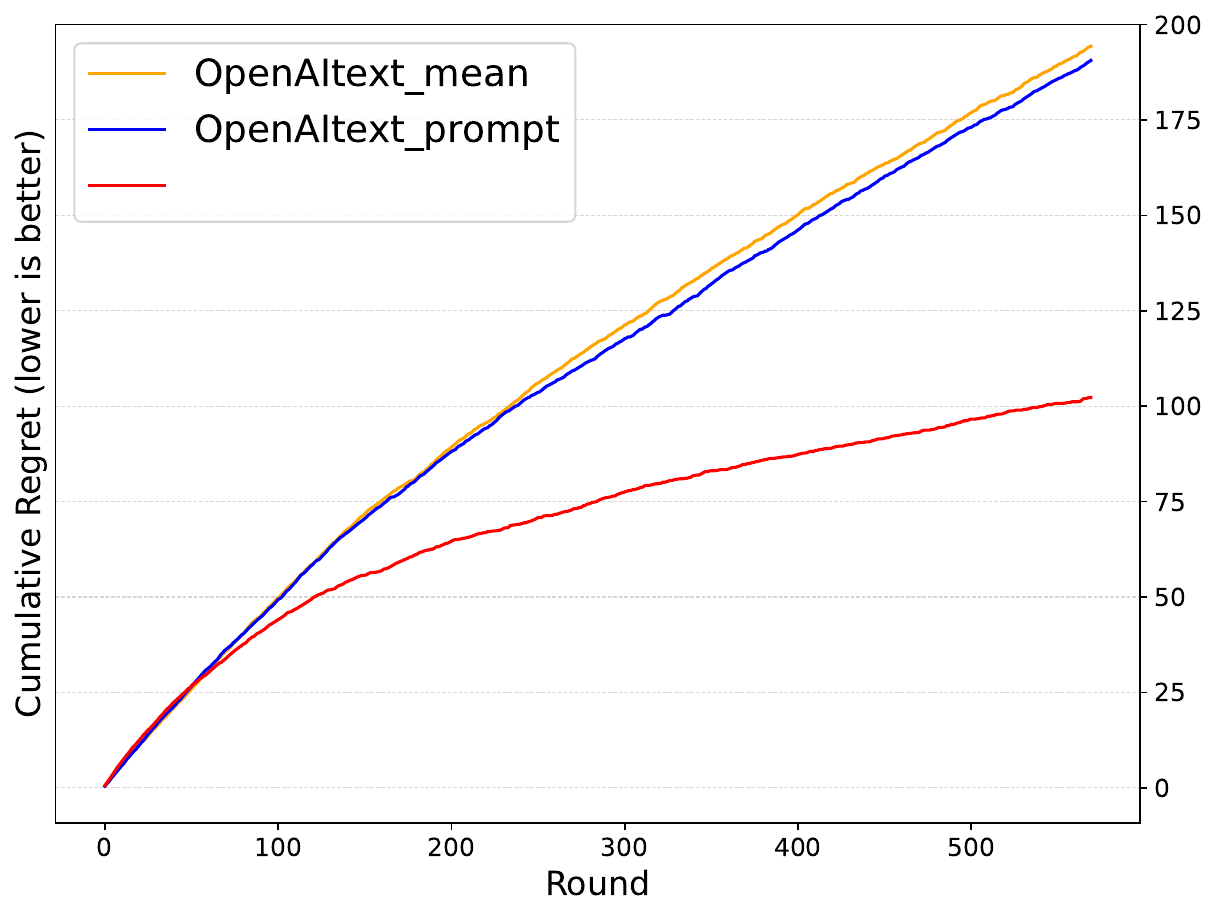}
    \vspace{-8mm}
    \end{center}
    \caption{Failed versus successful examples.}
    \label{fig:failure_exs}
\end{minipage}
\vspace{-4mm}
\end{wrapfigure}
This subsection highlights the practical challenges involved in designing an effective feature function. 
We construct synthetic simulations based on the MMLU dataset \citep{hendrycks2021mmlu}. 
To this end, we use OpenAI's \texttt{text-embedding-3-large} model to implement two straightforward embedding methods: \texttt{OpenAItext\_prompt}, which uses prompting, and \texttt{OpenAItext\_mean}, which relies on averaging embeddings. 
Their experimental details are defer to App.~\ref{sec:mmlu_experimental_details}

From Fig.~\ref{fig:failure_exs}, we see that the slopes of \texttt{OpenAItext\_mean} and \texttt{OpenAItext\_prompt} almost do not change with rounds. 
This means the regret keeps accumulating, and hence the learning does not progress a lot.
In contrast, the red curve resembles the learning behavior we want.
The slope of the red curve reduces as the rounds increase. 
This results in small cumulative regret, reflecting that the learning agent is making fewer and fewer mispredictions and converging the behavior toward the best routing policy. 
In the next subsection, we propose a generic strategy to construct model embeddings that aims to achieve behavior similar to the red curve, and we test its implementations on real-world datasets in experiments.
% In the next subsection, we propose a generic strategy to construct model embeddings achieving a similar behavior to the red curve and then test its realizations on real-world datasets in experiments.

\vspace{-2mm}\subsection{The Proposed Method}
\label{sec:proposed_model_emb}

Note that by design, $\phi(x_t, a_k)$ combines the information from the user side, $x_t$, and the information from the LLM side, $a_k$.
Since the text models we selected are well-recognized sentence encoders, the failure cases discussed in \S~\ref{sec:failure_cases} are most likely due to careless design of the model embeddings.
% Therefore, it is imperative to ensure that $a_k$ accurately captures the characteristics of the corresponding LLM $k$.
Therefore, it is imperative to ensure that $a_k$ accurately captures the connection between the prospective queries and the LLM's expertise.
In this section, we introduce \emph{Category-Calibrated Fine-Tuning (CCFT)}, a generic strategy for constructing $a_k$ through contrastive fine-tuning combined with categorical weighting.

% The failure cases in \S~\ref{sec:failure_cases} show that a careful design of the model embedding $a$ is needed.
% In this section, we propose Category-Calibrated Fine-Tuning (CCFT), a generic strategy to train $a$ via contrastive fine-tuning and categorical weighting.
% Note that by design, $\phi(x_t, a)$ combines the information from the user side, $x_t$, and the information from the LLM side, $a$.  
% Therefore, it is imperative to make sure that $a_k$ captures the characteristics of LLM $k$.
Suppose the queries can be divided into $M$ categories.
% \mycmt{(s)}
% Suppose the questions can be divided into $M$ categories\footnote{A category can refer to a topic, a benchmark, or query type, depending on the dataset.}, such as fashion, opinion, history, biology, finance, and so on.
The 3D t-SNE plot of the failure examples in Fig.~\ref{fig:tsne-3d} shows that the text model tends to cluster queries from the same category.
It provides contrastive fine-tuning a good starting point.
Thus, we fine-tune the text model using a small offline query set that is disjoint from the online testing set.
Then, for each category $m$, we compute the category embedding $\xi_m$ by averaging the embeddings of offline queries belonging to that category, as generated by the fine-tuned text model. 
Note that the category embedding $\xi_m$ is not the model embedding $a_k$, for which an additional step is required, as described next.

To capture the unique characteristics of each LLM, we assume that every LLM is associated with a distinct Kiviat diagram, representing its areas of expertise.
Under this assumption, a natural approach is to compute a weighted combination of the category embeddings, where the weights are derived from the LLM’s scores on its Kiviat diagram. We refer to this mechanism as ``categorical weighting''.
In particular, we propose the following four weighting methods.
Denote $M$ category embeddings $(\xi_1, \xi_2, \ldots, \xi_M) := \xi$.
For each LLM, let $(s_{k,1}, s_{k,2}, \ldots, s_{k,M})^{\top} := s_k$ be the score vector over the categories.
The first weighting method, coined ``\texttt{perf}'', defines the model embedding as
\eqary{
    a_k = \xi \operatorname{softmax}(s_k) , 
    \num\label{eq:def_perf}
}
in which $s_{k,m}$ is the model performance on category $m$.
If the score $s_{k,m}$ is a function of model performance and model cost, we coin the resulting $a_k$ ``\texttt{perf\_cost}''.
Note that \texttt{perf} and \texttt{perf\_cost} take all score values into account.
In reality, a common scenario involves a few strong models that specialize in different categories, alongside several weaker models that perform comparably within specific domains.
In such a case, we could weight an LLM only on categories it is good at, and leave the other categories handled by other LLMs.
Formalizing this idea, we propose ``\texttt{excel\_perf\_cost}'' as
\eqary{
    a_k = \xi \operatorname{softmax}\left(\operatorname{top}^{(\tau)}(s_k) \right),
    \num\label{eq:def_excel_perf_cost}
}
and ``\texttt{excel\_mask}'' as
\eqary{
    a_k = \xi \frac{\operatorname{mask}^{(\tau)}(s_k)}{\tau}.
    \num\label{eq:def_excel_mask}
}
Let $\tau \in \{1, 2, \ldots, K \}$, and let $s_{(\tau),m}$ denote the $\tau$-th largest value in $\{ s_{1,m}, s_{2,m}, \ldots, s_{K,m} \}$\footnote{Sorting $\{ s_{1,m}, s_{2,m}, \ldots, s_{K,m} \}$ results in $s_{(1),m} \geq s_{(2),m} \geq \ldots \geq s_{(K),m}$, and $s_{(\tau),m}$ will be located at the $\tau$-th position.}.
Function $\operatorname{top}^{(\tau)}(s_k)$ produces a real-valued vector with $m$-th entry $\operatorname{top}^{(\tau)}(s_k)_{m} = s_{k,m} \mathbf{1}[s_{k,m} \geq s_{(\tau),m}]$.
Similarly, function $\operatorname{mask}^{(\tau)}(s_k)$ produces a binary vector with $m$-th entry $\operatorname{mask}^{(\tau)}(s_k)_{m} = \mathbf{1}[s_{k,m} \geq s_{(\tau),m}]$.
% Let $\tau \in [M]$, and let $s_{k,(\tau)}$ denote the $\tau$-th largest value in $s_k$. 
% Function $\operatorname{top}^{(\tau)}(s_k)$ produces a real-valued vector with $m$-th entry
% \eqary{
%     \operatorname{top}^{(\tau)}(s_k)_{m} = s_{k,m} \mathbf{1}[s_{k,m} \geq s_{(\tau),m}] = 
%     \begin{cases}
%         s_{k,m} & \text{if } s_{k,m} \geq s_{(\tau),m},\\
%         0 & \text{otherwise.}
%     \end{cases}
% }
% Function $\operatorname{mask}^{(\tau)}(s_k)$ produces a binary vector with $m$-th entry
% \eqary{
%     \operatorname{mask}^{(\tau)}(s_k)_{m} = \mathbf{1}[s_{k,m} \geq s_{(\tau),m}] = 
%     \begin{cases}
%         1 & \text{if } s_{k,m} \geq s_{(\tau),m},\\
%         0 & \text{otherwise.}
%     \end{cases}
% }
Now, we have proposed four variants for computing an LLM embedding.
% \mycmt{(s)}
% Now, we have four variants of computing an LLM embedding.
% The main idea behind these mechanisms is that the scores reflect the expertise of an LLM over the categories, the category embeddings abstract the prospective questions, and the definitions $a_k$ show how we model the connection of these two key factors.

\paragraph{Categorical Weighting without Scores}

Note that the weighting mechanisms (\ref{eq:def_perf}), (\ref{eq:def_excel_perf_cost}), and (\ref{eq:def_excel_mask}) require score information, which might not always be the case.
Next, we show that we can still perform a way of categorical weighting under a mild condition.

Suppose the label of a query records its best matching LLM.
Thus, we use $f_{km}$ to denote the proportion of label $k$ in category $m$.
We assume the offline data is generated as follows:
From each category $m$, a set of query embeddings $\mathcal{Q}_m$ of size $n$ is sampled.
The offline data is $\{ \mathcal{G}_k \}_{k=1}^{K}$, formed by regrouping $\{ \mathcal{Q}_m \}_{m=1}^{M}$ according to the labels.
Given the offline data, we propose to compute the model embedding, for each $k \in \{1, 2, \ldots, K\}$, as  
\eqary{
    a_k = \sum_{q \in \mathcal{G}_k} q / |\mathcal{G}_k|.
    \num\label{eq:def_mix_inst}
}
The following proposition justifies the proposed mechanism; the proof is deferred to Appendix~\ref{proof:weight_without_scores}.
\begin{proposition}
\label{thm:weight_without_scores}
    Let $f_{km}$, $\{ \mathcal{Q}_m \}_{m=1}^{M}$, and $\{ \mathcal{G}_k \}_{k=1}^{K}$ be defined as above.
    Let $\mathbb{E}[Q_m]$ denote the expected embedding of queries in category $m$.
    Assume the embedding distribution within category $m$ is independent of label $k$\footnote{This is a reasonable assumption, for instance, when preferences are determined by user traits.}.
    Then, for each $k \in \{1, 2, \ldots, K\}$, the average embedding (\ref{eq:def_mix_inst}) is an unbiased estimate for 
    $\sum_{m=1}^{M} \frac{f_{km}}{\sum_{j=1}^{M} f_{kj}} \mathbb{E}[Q_m]$.
    % \eqary{
    %     \sum_{m=1}^{M} \frac{f_{km}}{\sum_{j=1}^{M} f_{kj}} \mathbb{E}[Q_m].
    % }
\end{proposition}

Viewing $\mathbb{E}[Q_m]$ as $\xi_m$, the proposition shows that, even if there is no score information available, averaging over query embeddings still offers a way of categorical weighting in terms of label proportions: the weighting coefficients are $ \frac{f_{km}}{\sum_{j=1}^{M} f_{kj}}, m=1, \ldots, M$.
Note that a constraint is the constant $n$ over categories.
The constraint can be relaxed if we know the number of samples from each category.
Furthermore, when a pool of pairwise comparison results is gathered, one can build a rank over LLMs to determine the best matching model. 
Therefore, the $\{ \mathcal{G}_k \}$ setting and the proposed model embedding mechanism in this section fit naturally into many industrial applications.

\vspace{-1mm}\section{Experiments}
\label{sec:experiments}
% \mycmt{(FGTS.CDB and its SGLD module)}
We implemented FGTS.CDB shown in Algorithm~\ref{alg:fgts-cdb}. 
Sampling from the posterior $p^j(\theta \mid S_{t-1})$ in Step 5 is implemented by the Stochastic Gradient Langevin Dynamics (SGLD) algorithm \citep{welling2011GDLD}.
% \mycmt{(Text embedding models)}
Four text embedding models are chosen to generate embeddings for queries.
They are: OpenAI's \texttt{text-embedding-3-large} \citep{openai-text-embedding-3-large}, \texttt{all-MiniLM-L6-v2} \citep{sentence-transformers-all-MiniLM-L6-v2}, \texttt{all-mpnet-base-v2} \citep{sentence-transformers-all-mpnet-base-v2}, and \texttt{intfloat/e5-base} \citep{wang2022e5}.
In this paper, we refer to the models as \texttt{OpenAItext}, \texttt{MiniLM}, \texttt{mpnet}, and \texttt{e5b}, respectively.
\texttt{OpenAItext} serves as a strong and general-purpose embedding model.
We compare its embeddings with those generated by the fine-tuned \texttt{MiniLM}, \texttt{mpnet}, and \texttt{e5b} models to examine their impact on the feature function $\phi(x, a)$ derived from a text model.
% \mycmt{(Contrastive fine-tuning)}
The fine-tuning for an open-sourced text embedding model is implemented by contrastive learning \citep{Khosla20SCL, Nils2019sentenceBERT}. 
We first build similar and dissimilar query pairs according to their source category or benchmark.
Then, the cosine-similarity loss is used to fine-tune the model.
For instance, \texttt{eb5\_E4} means that \texttt{eb5} is fine-tuned for four epochs, and \texttt{mpnet\_E2} represents \texttt{mpnet} fine-tuned for two epochs.

\vspace{-2mm}\subsection{RouterBench}
\label{sec:exp_routerb}
RouterBench \citep{hu2024routerbench} is a comprehensive benchmark designed for evaluating LLM routing methods. 
It provides over 405k precomputed inference outputs from eleven diverse LLMs across seven tasks (MMLU, MT-Bench, MBPP, HellaSwag, Winogrande, GSM8K, ARC). 
The dataset includes detailed performance and cost metadata, enabling systematic analysis of routing strategies.
The metadata, organized into a table in \cite{hu2024routerbench} is included as Tab.~\ref{table:Table_1_routerb} in App.~\ref{sec:more_categorical_weighting}.

In the following, we describe how to use the metadata and the queries in each benchmark to construct an LLM embedding $a_k$.
The learning process is divided into two phases: an offline fine-tuning phase, during which the model embeddings are learned, and an online testing phase, where a realization of FGTS.CDB is evaluated through a sequence of queries. 
To apply the proposed method, we need to compute $s_k$ and $\xi$.
Each benchmark $m$ is treated as a distinct category. 
To compute the corresponding category embedding $\xi_m$, we apply a text embedding model to five queries sampled from benchmark $m$, and then take the average of their embeddings.
% \mycmt{(compare; distinctive) We only need 35 queries vs. MixLLM \citep{wang2025mixllm} needs 30\% of the data. $\Rightarrow$ See the end of this section.}
These sampled queries are excluded from the online learning phase to prevent data leakage.
Suppose \texttt{e5b\_E4} is applied to generate the embeddings. 
Then, we obtain $\xi^{(\texttt{e5b\_E4})} = \left( \xi_{\textrm{MMLU}}^{(\texttt{e5b\_E4})}, \xi_{\textrm{MT-Bench}}^{(\texttt{e5b\_E4})}, \ldots, \xi_{\textrm{ARC}}^{(\texttt{e5b\_E4})} \right)$.

\begin{table*}[htbp]
\centering
\caption{Scores of \texttt{Perf\_cost}, \texttt{Excel\_perf\_cost}, and \texttt{Excel\_mask}}
\label{table:weighting}
\begin{adjustbox}{max width=\textwidth}
\begin{tabular}{lcccccccccccccccccccccccc}
\toprule
\textbf{LLM} & 
\multicolumn{3}{c}{\textbf{MMLU}} & 
\multicolumn{3}{c}{\textbf{MT-Bench}} & 
\multicolumn{3}{c}{\textbf{MBPP}} & 
\multicolumn{3}{c}{\textbf{HellaSwag}} & 
\multicolumn{3}{c}{\textbf{Winogrande}} & 
\multicolumn{3}{c}{\textbf{GSM8k}} & 
\multicolumn{3}{c}{\textbf{ARC}} \\
\cmidrule(lr){2-4}
\cmidrule(lr){5-7}
\cmidrule(lr){8-10}
\cmidrule(lr){11-13}
\cmidrule(lr){14-16}
\cmidrule(lr){17-19}
\cmidrule(lr){20-22}
& (i) & (ii) & (iii) 
& (i) & (ii) & (iii) 
& (i) & (ii) & (iii) 
& (i) & (ii) & (iii) 
& (i) & (ii) & (iii) 
& (i) & (ii) & (iii) 
& (i) & (ii) & (iii) \\
% & (\ref{eq:def_perf}) & (\ref{eq:def_excel_perf_cost}) & (\ref{eq:def_excel_mask}) 
% & (\ref{eq:def_perf}) & (\ref{eq:def_excel_perf_cost}) & (\ref{eq:def_excel_mask}) 
% & (\ref{eq:def_perf}) & (\ref{eq:def_excel_perf_cost}) & (\ref{eq:def_excel_mask}) 
% & (\ref{eq:def_perf}) & (\ref{eq:def_excel_perf_cost}) & (\ref{eq:def_excel_mask}) 
% & (\ref{eq:def_perf}) & (\ref{eq:def_excel_perf_cost}) & (\ref{eq:def_excel_mask}) 
% & (\ref{eq:def_perf}) & (\ref{eq:def_excel_perf_cost}) & (\ref{eq:def_excel_mask}) 
% & (\ref{eq:def_perf}) & (\ref{eq:def_excel_perf_cost}) & (\ref{eq:def_excel_mask}) \\
\midrule
WizardLM 13B     & 0.562 & 0 & 0 & 0.796 & 0 & 0 & 0.363 & 0 & 0 & 0.600 & 0 & 0 & 0.510 & 0 & 0 & 0.492 & 0 & 0 & 0.657 & 0 & 0 \\
Mistral 7B       & 0.558 & 0 & 0 & 0.779 & 0 & 0 & 0.349 & 0 & 0 & 0.517 & 0 & 0 & 0.561 & 0 & 0 & 0.399 & 0 & 0 & 0.640 & 0 & 0 \\
Mixtral 8x7B     & 0.721 & 0.721 & 1 & 0.920 & 0.920 & 1 & 0.572 & 0.572 & 1 & 0.634 & 0 & 0 & 0.673 & 0.673 & 1 & 0.485 & 0 & 0 & 0.837 & 0.837 & 1 \\
Code Llama 34B   & 0.553 & 0 & 0 & 0.795 & 0 & 0 & 0.464 & 0 & 0 & 0.431 & 0 & 0 & 0.612 & 0 & 0 & 0.424 & 0 & 0 & 0.635 & 0 & 0 \\
Yi 34B           & 0.727 & 0.727 & 1 & 0.937 & 0.937 & 1 & 0.331 & 0 & 0 & 0.834 & 0.834 & 1 & 0.743 & 0.743 & 1 & 0.509 & 0.509 & 1 & 0.873 & 0.873 & 1 \\
GPT-3.5          & 0.700 & 0.700 & 1 & 0.907 & 0.907 & 1 & 0.649 & 0.649 & 1 & 0.695 & 0.695 & 1 & 0.623 & 0.623 & 1 & 0.543 & 0.543 & 1 & 0.844 & 0.844 & 1 \\
Claude Instant V1& 0.368 & 0 & 0 & 0.862 & 0 & 0 & 0.547 & 0 & 0 & 0.704 & 0.704 & 1 & 0.507 & 0 & 0 & 0.561 & 0.561 & 1 & 0.812 & 0 & 0 \\
Llama 70B        & 0.629 & 0 & 0 & 0.853 & 0 & 0 & 0.300 & 0 & 0 & 0.627 & 0 & 0 & 0.498 & 0 & 0 & 0.486 & 0 & 0 & 0.784 & 0 & 0 \\
Claude V1        & 0.312 & 0 & 0 & 0.920 & 0.920 & 1 & 0.497 & 0 & 0 & -0.131 & 0 & 0 & 0.516 & 0 & 0 & 0.099 & 0 & 0 & 0.798 & 0 & 0 \\
Claude V2        & 0.456 & 0 & 0 & 0.840 & 0 & 0 & 0.567 & 0.567 & 1 & -0.554 & 0 & 0 & 0.392 & 0 & 0 & -0.011 & 0 & 0 & 0.454 & 0 & 0 \\
\bottomrule
\end{tabular}
\end{adjustbox}
\end{table*}

We then use the metadata in Tab.~\ref{table:Table_1_routerb} and the induced Tab.~\ref{table:weighting} to calculate the scores $s_k$.
The most straightforward way is to take only the performance columns in Tab.~\ref{table:Table_1_routerb}, which corresponds to \texttt{Perf}.
For instance, taking the performance values in the fifth row of Tab.~\ref{table:Table_1_routerb}, we obtain $s_{\textrm{Yi 34B}}^{(\texttt{Perf})} = (0.743, 0.938, \ldots, 0.882)$.
We also implement three other scoring ways, \texttt{Perf\_cost}, \texttt{Excel\_perf\_cost}, and \texttt{Excel\_mask}. 
Using the Perf and the Cost columns in Tab.~\ref{table:Table_1_routerb}, scores of \texttt{Perf\_cost} is calculated by $\textrm{Perf} - \lambda \textrm{Cost}$ with $\lambda = 0.05$ being a balance parameter.
The resulting values are listed in the columns of Tab.~\ref{table:weighting} indexed by (i).
With \texttt{Perf\_cost} in hand, \texttt{Excel\_perf\_cost} keeps the original \texttt{Perf\_cost} score if it is ranked as the top-$\tau$ in the column and assigns $0$ otherwise. 
Here we choose $\tau=3$.
The \texttt{Excel\_perf\_cost} scores are listed under index (ii).
\texttt{Excel\_mask} further masks the nonzero values of \texttt{Excel\_perf\_cost} as $1$ and lists them under (iii) in Tab.~\ref{table:weighting}. 
Finally, we feed the scores $s_k$ and the model embeddings $\xi$ to (\ref{eq:def_perf}), (\ref{eq:def_excel_perf_cost}), and (\ref{eq:def_excel_mask}) to obtain the model embedding $a_k$
% \mycmt{(model embedding, LLM embedding, LLM's embedding?)}.
Specifically, \texttt{Perf} and \texttt{Perf\_cost} are fed to (\ref{eq:def_perf}), \texttt{Excel\_perf\_cost} to (\ref{eq:def_excel_perf_cost}), and \texttt{Excel\_mask} to (\ref{eq:def_excel_mask}).
Using the above symbols, we can denote, for instance, the model embedding of Yi 34B calculated via text embedding model \texttt{e5b\_E4}, scoring method \texttt{Perf\_cost}, and weighting equation (\ref{eq:def_perf}) as 
$
a_{\textrm{Yi 34B}}^{(\texttt{e5b\_E4\_}\texttt{Perf\_cost})} = \xi^{(\texttt{e5b\_E4})} \operatorname{softmax}\left(s_{\textrm{Yi 34B}}^{(\texttt{Perf\_cost})}\right)\footnote{Note that \texttt{e5b}, \texttt{mpnet}, and \texttt{MiniLM} are text models that generate embeddings. They need to be distinguished from the LLMs, listed in the leftmost column of Tab.~\ref{table:weighting}, which are candidates selected to generate responses in online learning.}.
$ 
In addition, we append all $14$ metadata (Perf and Cost over seven benchmarks) of an LLM at the end of its embedding.
Finally, the feature function $\phi(x, a_k)$ is computed as the normalized Hadamard product $x * a_{k}$. 
To compare the effectiveness of fine-tuning, we use a suffix to indicate whether a fine-tuned model or an original model generates an embedding. 
For instance, the string ``\texttt{e5b\_E4\_Excel\_perf\_cost\_exp}'' means ``belonging to the experimental group, each embedding is generated by \texttt{e5b} fine-tuned with four epochs with the \texttt{Excel\_perf\_cost} weighting mechanism'', and ``\texttt{e5b\_E4\_Excel\_perf\_cost\_ctrl}'' means ``belonging to the control group, each embedding is generated by the original \texttt{e5b} with the \texttt{Excel\_perf\_cost} weighting mechanism.
\new
Since we cannot fine-tune \texttt{OpenAItext}, we use handcraft prompts to generate LLM embeddings. 
We plug the offline query samples and the metadata of the dataset into the prompt (Listing~\ref{code_snippet:prompt_routerbench} in App.~\ref{sec:prompts}) to get the model description, and then feed it to \texttt{OpenAItext} to obtain a model embedding.
We use performance metadata as the utility function, from which we generate online feedback via the BLT protocol and compute regret at each round.
% We use performance metadata as the utility function, generating online feedback via the BLT protocol and computing regret at each round.

\begin{figure}[htbp]
    \centering
    \begin{minipage}{0.24\textwidth}
        \centering
        \includegraphics[width=\linewidth]{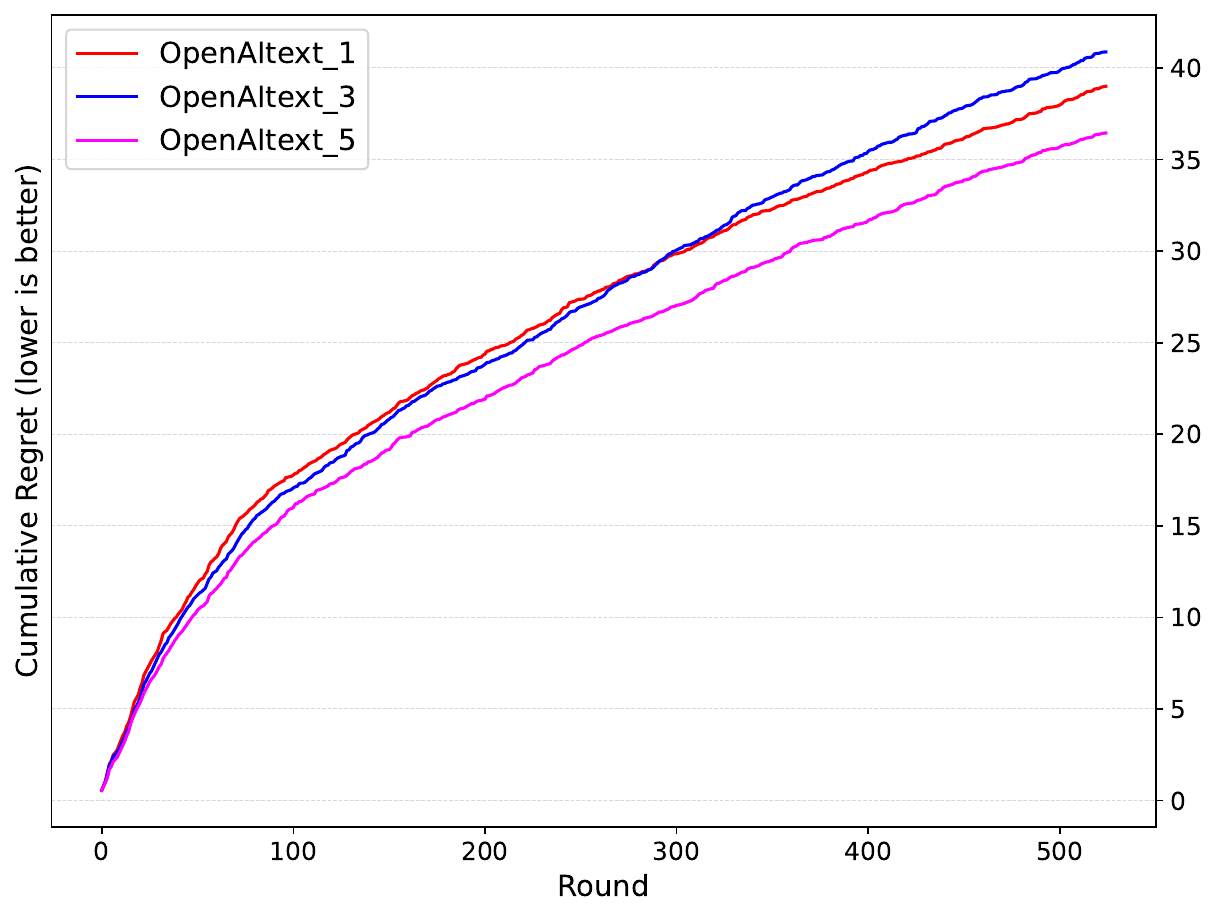}
        \subcaption{\texttt{OpenAItext} results}
        \label{fig:rtb-main-a}
    \end{minipage}
    \hfill
    \begin{minipage}{0.24\textwidth}
        \centering
        \includegraphics[width=\linewidth]{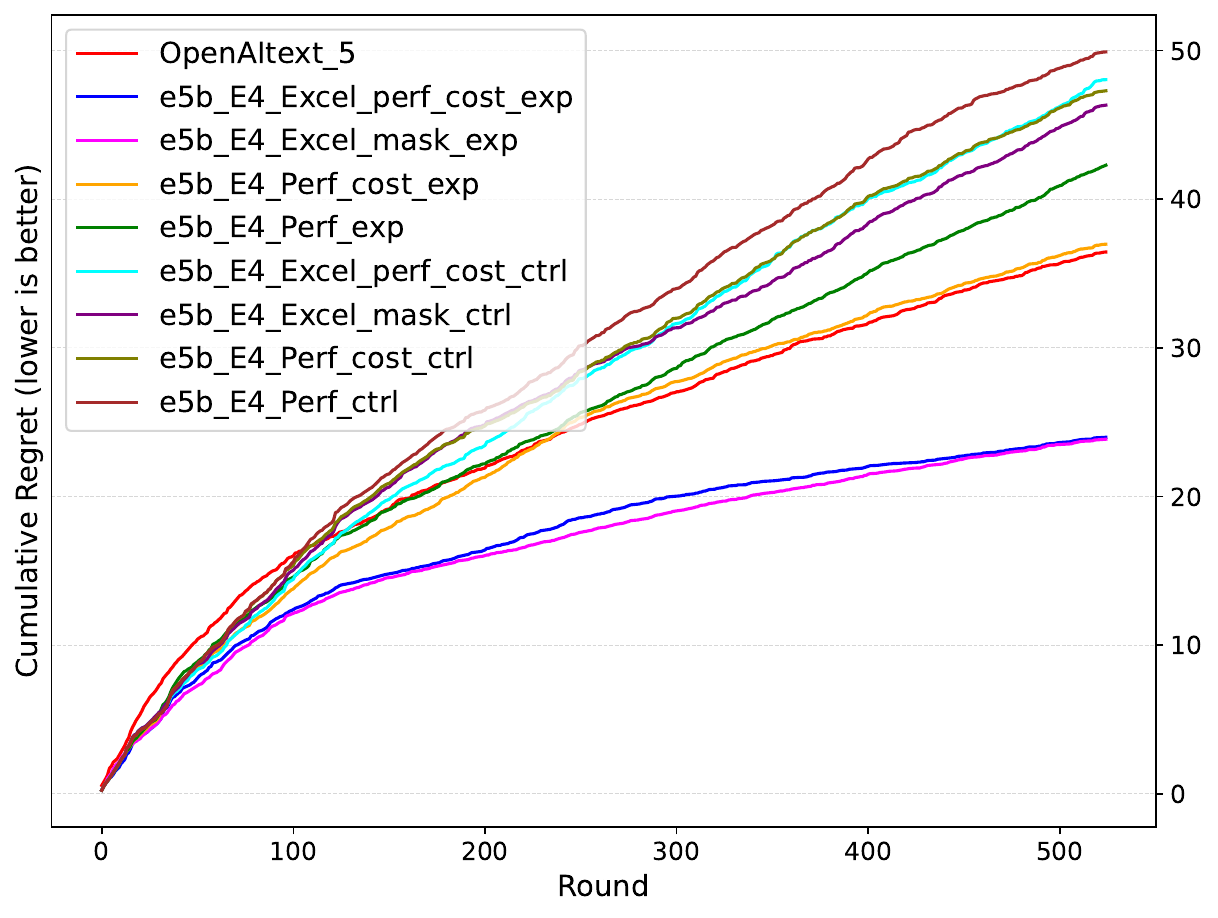}
        \subcaption{\texttt{e5b\_E4} results}
        \label{fig:rtb-main-b}
    \end{minipage}
    \hfill
    \begin{minipage}{0.24\textwidth}
        \centering
        \includegraphics[width=\linewidth]{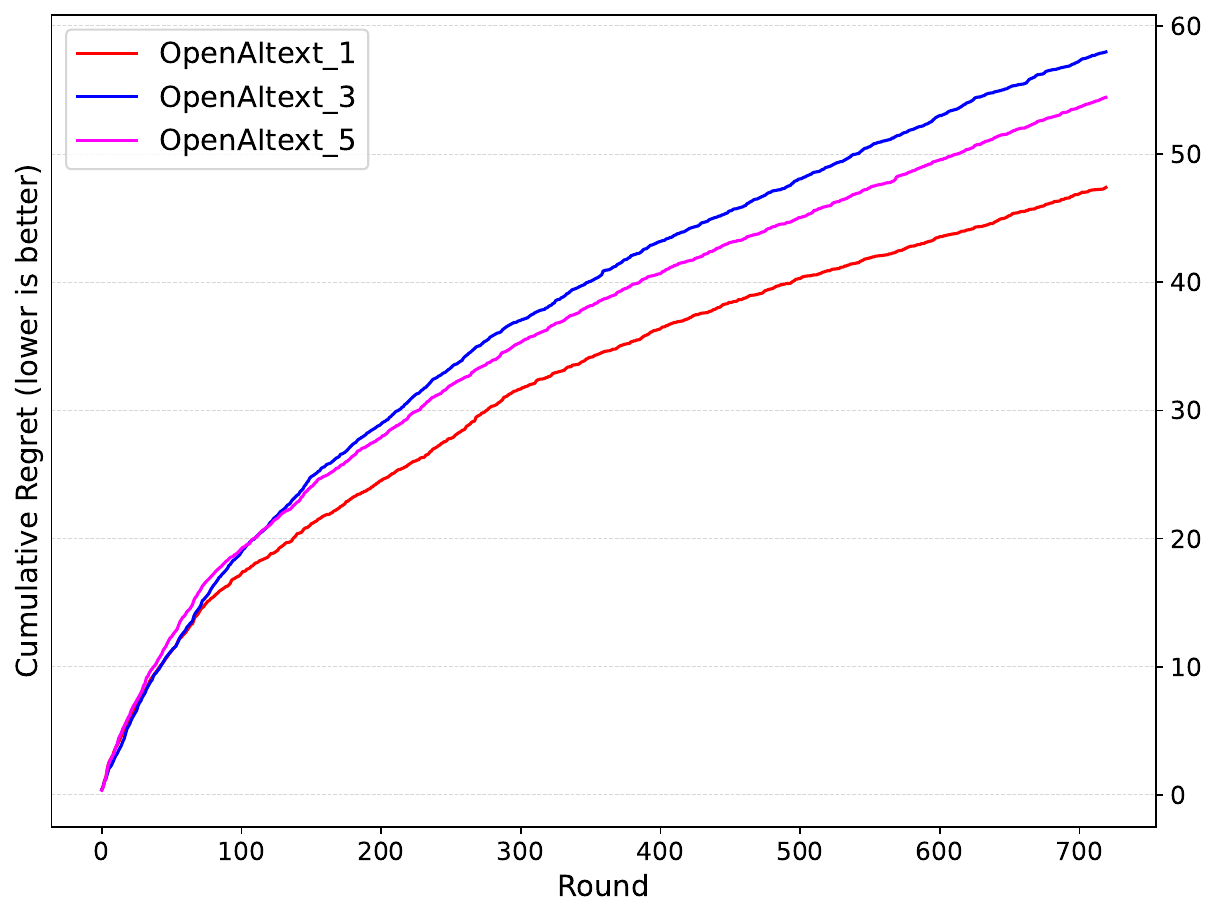}
        \subcaption{\texttt{OpenAItext} results}
        \label{fig:rbst-main-a}
    \end{minipage}
    \hfill
    \begin{minipage}{0.24\textwidth}
        \centering
        \includegraphics[width=\linewidth]{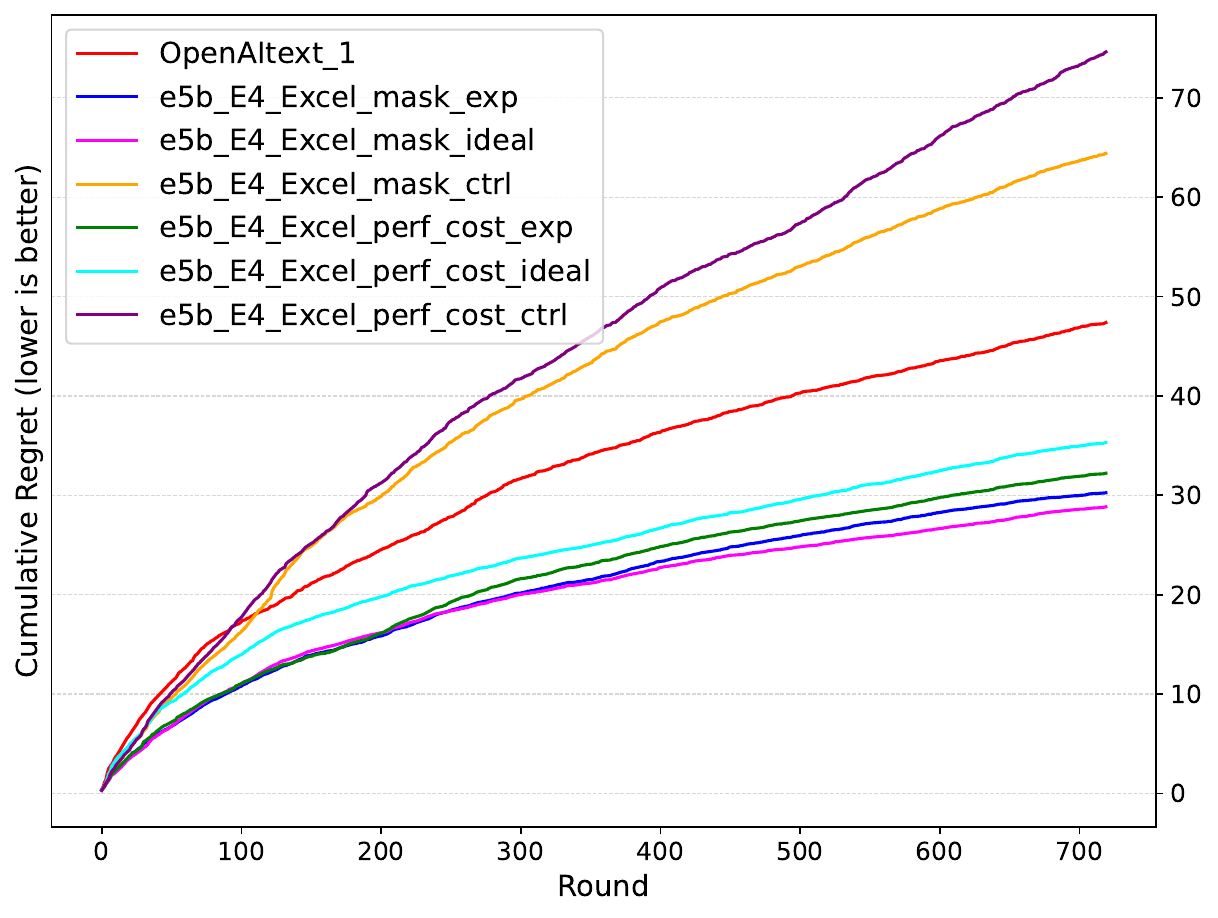}
        \subcaption{\texttt{e5b\_E4} results}
        \label{fig:rbst-main-b}
    \end{minipage}
    \vspace{-3mm}
    \caption{Regret curves for RouterBench (a, b) and robust generalization (c, d).}
    \label{fig:combined-rtb-rbst}
    \vspace{-3mm}
\end{figure}

We run \texttt{OpenAItext} with one, three, and five queries, each for five runs, and average the cumulative regrets to plot three curves\footnote{Unless mentioned, each regret curve reported is the average of 5 runs.} in Fig.~\ref{fig:rtb-main-a}.
% \mycmt{Is it 1, 3, and 5 or one, three, and five?}
For open-sourced embdeeing models, we fine-tune to obtain \texttt{e5b\_E2}, \texttt{e5b\_E4}, \texttt{mpnet\_E2}, \texttt{mpnet\_E4}, \texttt{MiniLM\_E2}, and \texttt{MiniLM\_E4}.
For each embedding model, four weighting mechanisms (\texttt{Perf}, \texttt{Perf\_cost}, \texttt{Excel\_perf\_cost}, \texttt{Excel\_mask}) are implemented.
The corresponding control group also undergoes the online testing phase.
There are a total of 8 curves for one embedding model.
The best performed \texttt{e5b\_E4} is shown in Fig.~\ref{fig:rtb-main-b}. 
Results for all models can be found in Fig.~\ref{fig:grid-rtb-models-all}, App.~\ref{sec:more_routerb_results}.

% --- 1x4 layout with minipage ---

We obtain the following observations from the regret curves.
In Fig.~\ref{fig:rtb-main-a}, the number of queries in the prompt does not affect \texttt{OpenAItext} too much.
Thus, we choose the best version, \texttt{OpenAItext\_5}, to compare with the results generated by our proposed CCFT in Fig.~\ref{fig:rtb-main-b}.
First, the experimental group outperforms the control group, showing the advantage of fine-tuning.
Second, \texttt{e5b\_E4\_Excel\_perf\_cost\_exp} and \texttt{e5b\_E4\_Excel\_mask\_exp} outperform \texttt{OpenAItext\_5}, showing that even an open-sourced model, with a careful design, can generate better embeddings than a strong general-purpose model. 
Third, when examining all plots in Fig.~\ref{fig:grid-rtb-models-all-b} -- \ref{fig:grid-rtb-models-all-f}, the previous two observations hold for all other embedding models, showing convincing evidence for the effectiveness of CCFT we proposed.
Fourth, by comparing \texttt{e5b\_E4\_Excel\_perf\_cost\_exp} with \texttt{e5b\_E4\_Perf\_cost\_exp}, we observe that weighting via \texttt{Excel\_perf\_cost} yields better performance than weighting via \texttt{Perf\_cost}.
% \mycmt{We compare with $a^*_{t}$, not just $a^*$.}
This suggests that, instead of weighting over all categories, it is more effective to weight only the categories in which the LLM demonstrates expertise.
In addition to achieving lower accumulated regret, both \texttt{Excel\_perf\_cost} and \texttt{Excel\_mask} incorporate a performance-cost trade-off, enhancing their practical applicability and flexibility through the tunable parameter $\lambda$.
App.~\ref{sec:more_routerb_compare} presents a comparison between our approach and the closely related MixLLM \citep{wang2025mixllm}.

\vspace{-2mm}\subsubsection{Generalization to an Unseen Benchmark}
\label{sec:robust_gen}
% --- 1x2 layout wrapfigure ---
% \begin{wrapfigure}{r}{0.6\textwidth}
% \vspace{-4mm}
% \begin{minipage}{\linewidth}
%     \centering
%     \begin{subfigure}{0.48\linewidth}
%         \centering
%         \includegraphics[width=\linewidth]{plots/routerb_rbst2_m42_set4_figure_0.pdf}
%         \vspace{-6mm}
%         \caption{\texttt{OpenAItext} results}
%         \label{fig:rbst-main-a}
%     \end{subfigure}
%     \begin{subfigure}{0.48\linewidth}
%         \centering
%         \includegraphics[width=\linewidth]{plots/routerb_rbst1_m42_set1_figure_0.pdf}
%         \vspace{-6mm}
%         \caption{\texttt{e5b\_E4} results}
%         \label{fig:rbst-main-b}
%     \end{subfigure}
    
%     \vspace{-2mm}
%     \caption{Regret curves for robust generalization.}
%     \label{fig:rbst-main}
%     \vspace{-4mm}
% \end{minipage}
% \end{wrapfigure}
Although an online learning setting evaluates an algorithm’s adaptivity by providing sequentially and randomly shuffled inputs, the algorithm may still access metadata from all benchmarks during the offline stage.
To more rigorously assess adaptivity to an unseen benchmark, we modify the data-generation pipeline to ensure that one benchmark remains completely hidden from the algorithm throughout both the offline and online phases.
The queries and metadata of MT-Bench are removed, as its dataset is not large enough to induce a distribution shift scenario in the coming experiment.
For the remaining six benchmarks, the metadata for ARC is removed from Tab.~\ref{table:Table_1_routerb}, ensuring that the algorithm is oblivious to it from the outset.
An online learning sequence is composed of two sections.
First, we sample 60 queries from each benchmark, excluding ARC (i.e., five benchmarks in total), resulting in 300 queries. 
These are randomly shuffled to form the first section of the sequence.
Next, for the second section, we sample 120 queries from ARC and an additional 300 non-overlapping queries from the other five benchmarks, following the same procedure as in the first section. These 420 (i.e., $120 + 300$) queries are then shuffled to form the second section of the learning sequence.
This setup introduces a shift in the query distribution during the second section of online learning, as queries from a previously unseen benchmark are added to evaluate the robust generalization capability of the proposed method.
Due to the modification in how metadata is accessed, we introduce an additional suffix, \texttt{ideal}, which allows the model to access ARC's metadata.
Although the \texttt{ideal} case is not realistic in practice, comparing results from configurations ending with \texttt{ideal} and those ending with \texttt{exp} enables us to assess the adaptivity strength of our method.
For the rest of the experimental setting, we follow the same fine-tuning protocol in the last section to generate embeddings.
Based on the observations from figures~\ref{fig:rtb-main-b} and~\ref{fig:grid-rtb-models-all}, we implement the \texttt{Excel\_perf\_cost} and \texttt{Excel\_mask} weighting schemes due to their consistently strong performance.

% % --- 1x2 layout wrapfigure ---
% \begin{figure}[t]
%   \centering
%   \begin{subfigure}{0.48\textwidth}
%     \centering
%     \includegraphics[width=\linewidth]{plots/routerb_rbst2_m42_set4_figure_0.pdf}
%     \caption{OpenAI variants}
%     \label{fig:rbst-main-a}
%   \end{subfigure}\hfill
%   \begin{subfigure}{0.48\textwidth}
%     \centering
%     \includegraphics[width=\linewidth]{plots/routerb_rbst1_m42_set1_figure_0.pdf}
%     \caption{E5b variants}
%     \label{fig:rbst-main-b}
%   \end{subfigure}

%   \caption{Robust generalization: Cumulative regret curves (lower is better).}
%   \label{fig:rbst-main}
% \end{figure}

The results of \texttt{OpenAItext} and \texttt{e5b} are shown in figures~\ref{fig:rbst-main-a} and \ref{fig:rbst-main-b}.
Results for \texttt{mpnet} and \texttt{MiniLM} can be found in Fig.~\ref{fig:rtb-rbst-all} from App.~\ref{sec:more_routerb_results}.
We choose \texttt{OpenAItext\_1} to compare with the results generated by \texttt{e5b\_E4} according to Fig.~\ref{fig:rbst-main-a}.
First, obviously, the fine-tuning group outperforms the control group.
Second, \texttt{e5b\_E4\_Excel\_perf\_cost\_exp} and \texttt{e5b\_E4\_Excel\_mask\_exp}, the fine-tuning results via our CCFT strategy outperform that of \texttt{OpenAItext\_1}. 
% \mycmt{(revise; paper-wide) \texttt{OpenAItext\_1}.}
The case in the second observation holds for Fig.~\ref{fig:rtb-rbst-all-b} through Fig.~\ref{fig:rtb-rbst-all-f}, justifying the effectiveness of the proposed method.
% \mycmt{However, due to the unseen benchmark, the curve gaps between \texttt{OpenAItext\_1} become closer (REALLY?) to that in Fig.~\ref{fig:rtb-main-b}. I need to plot a figure to compare. Maybe openai\_0 suffers more in unseen benchmark.} 
Third, interestingly, we find that it is not always the case that an \texttt{ideal} curve outperforms the corresponding \texttt{exp} curve.
This can be found by comparing the pair (\texttt{e5b\_E4\_Excel\_perf\_cost\_exp}, \texttt{e5b\_E4\_Excel\_perf\_cost\_ideal}) with the pair (\texttt{e5b\_E4\_Excel\_mask\_exp}, \texttt{e5b\_E4\_Excel\_mask\_ideal}).
The situation also can be found in Fig.~\ref{fig:rtb-rbst-all}.
% \mycmt{In particular, only 3 pairs in Fig.~\ref{fig:rtb-rbst-all-a}, \ref{fig:rtb-rbst-all-e}, and \ref{fig:rtb-rbst-all-f} have \texttt{ideal} being better than \texttt{exp}.}
This phenomenon suggests weighting less may be better than weighting more, and it may be related to the last observation we have in the original RouterBench experiments that weighting over all benchmark embeddings is not a good idea.
Maybe the metadata alone is not enough
% for us
to make a weighting judgment; we might need to look into other aspects of the benchmarks in future work.

\vspace{-2mm}\subsection{MixInstruct}
\label{sec:exp_mixInst}
MixInstruct \citep{jiang2023mixinst_blender} is a 110K-example instruction-following benchmark built to evaluate LLM routing methods. 
It mixes data from four sources (Alpaca-GPT4, Dolly-15K, GPT4All-LAION, ShareGPT) with a 100k/5k/5k train/dev/test split. 
The authors run eleven popular open-source LLMs on the full set, then derive oracle pairwise preferences by prompting ChatGPT to compare every candidate pair per example.

\begin{table}[h!]
\caption{Models Ranked First by Percentage of Examples}
\label{table:mix-inst-top-percentage}
\centering
\begin{adjustbox}{max width=\textwidth}
\begin{tabular}{lccccccccccc}
\toprule
\textbf{Model} & Vicuna & MOSS & Open Assistant & Alpaca & Baize & ChatGLM & MPT & Koala & Dolly V2 & StableLM & FLAN-T5 \\
\midrule
\textbf{Percentage (\%)} & 21.22 & 12.91 & 12.61 & 11.61 & 11.61 & 8.51 & 7.61 & 6.71 & 4.50 & 1.90 & 0.80 \\
\bottomrule
\end{tabular}
\end{adjustbox}
\end{table}
% \mycmt{MixInstruct table is generated by \url{https://chatgpt.com/c/68cc1655-c370-8322-8021-c70b7cb8ea78}.}

Tab.~\ref{table:mix-inst-top-percentage}, adapted from Figure 1 of \cite{jiang2023mixinst_blender}, underscores the importance of LLM routing: selecting the best-matching LLM for each query is crucial, as any fixed-LLM strategy will yield no more than 22\% accuracy.
A distinctive characteristic of the MixInstruct dataset is the absence of an explicit category label, rendering the proposed methods in (\ref{eq:def_perf}), (\ref{eq:def_excel_perf_cost}), and (\ref{eq:def_excel_mask}) infeasible.
Since this challenge can naturally arise in practical settings, we use MixInstruct to evaluate our alternative formulation presented in (\ref{eq:def_mix_inst}), thereby validating its applicability under such a realistic constraint.
Moreover, the pairwise comparison labels in MixInstruct make it closely resemble datasets commonly encountered in industrial applications.

\begin{wrapfigure}{r}{0.6\textwidth}
\vspace{-4mm}
\begin{minipage}{\linewidth}
    \centering
    \begin{subfigure}{0.48\linewidth}
        \centering
        \includegraphics[width=\linewidth]{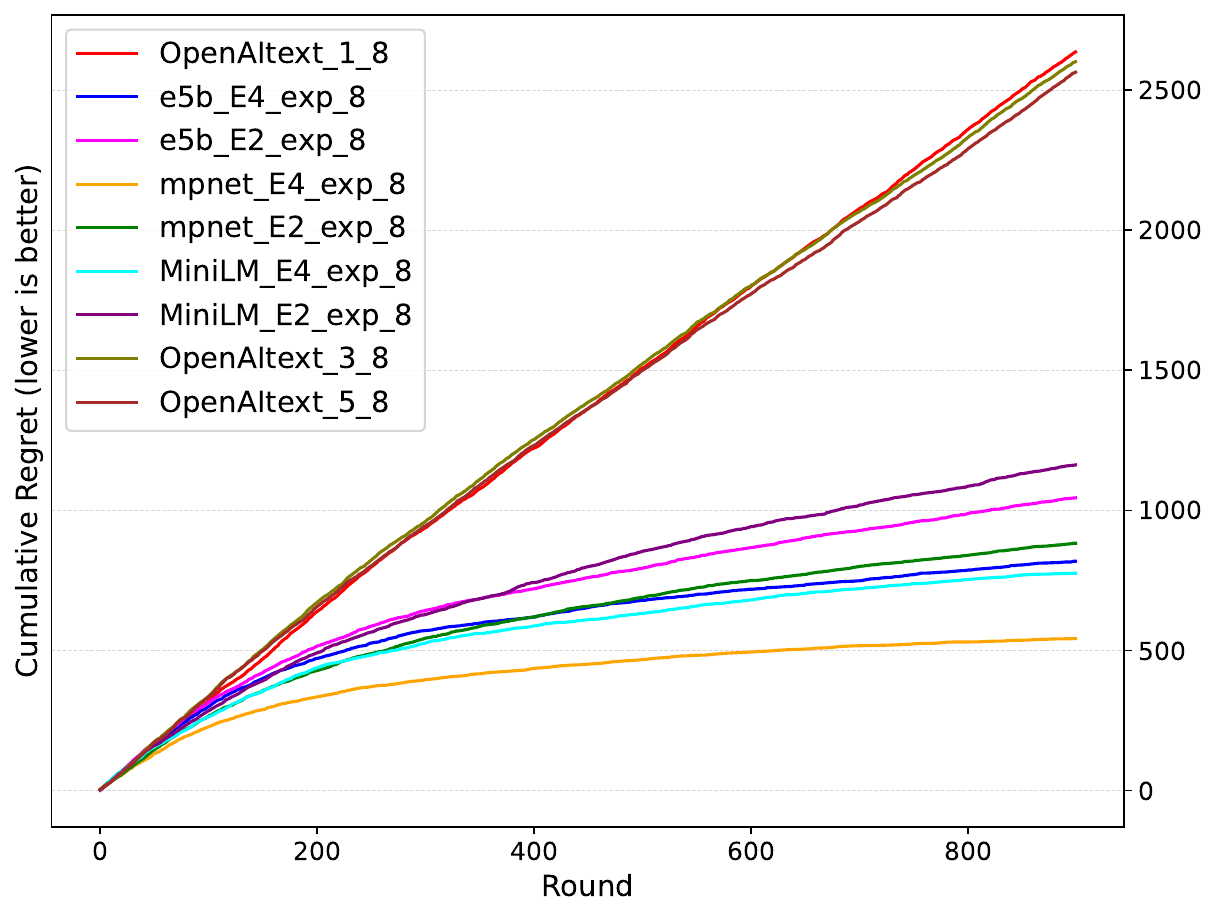}
        \vspace{-6mm}
        \caption{(\ref{eq:def_mix_inst}) versus \texttt{OpenAItext}.}
    \label{fig:mixinst-main-8}
    \end{subfigure}
    \begin{subfigure}{0.48\linewidth}
        \centering
        \includegraphics[width=\linewidth]{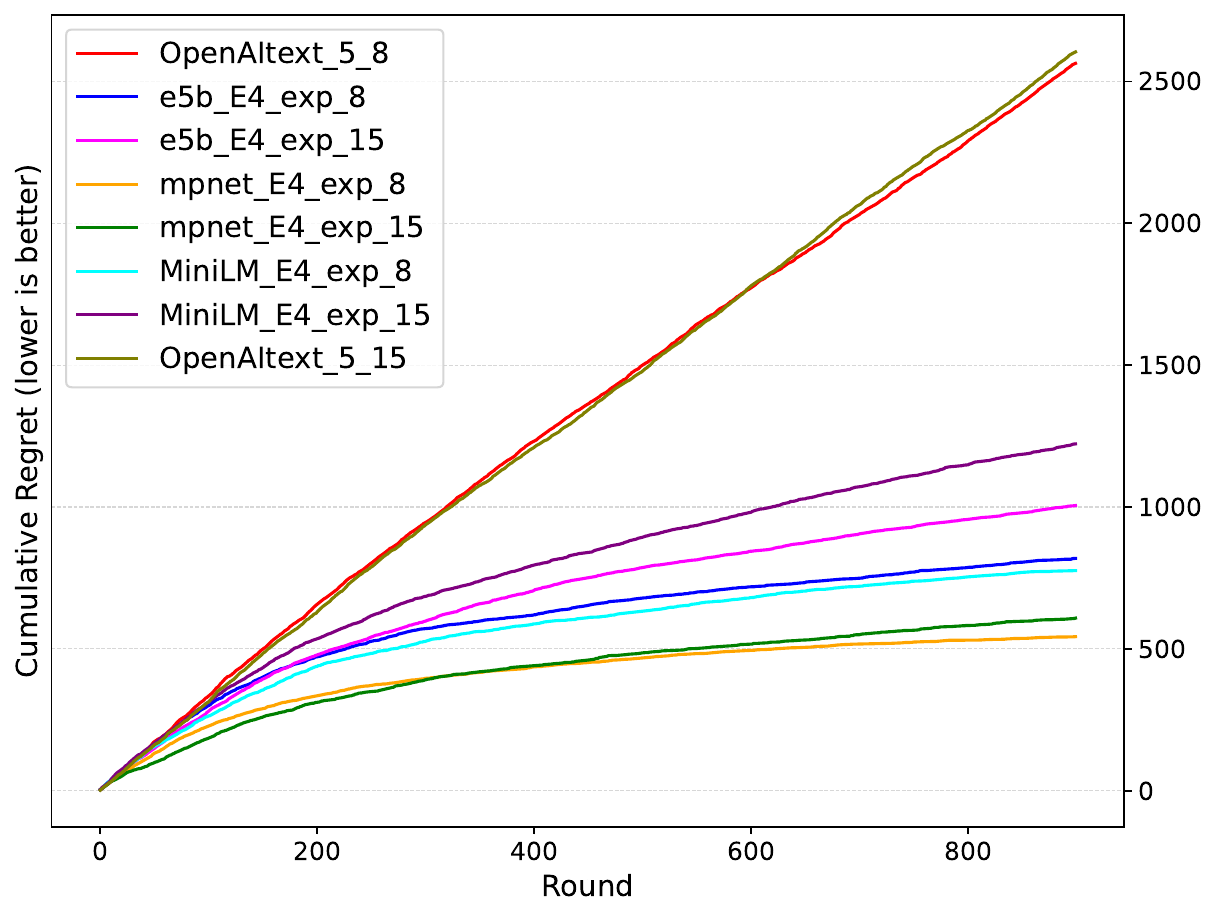}
        \vspace{-6mm}
        \caption{Ambiguity removal.}
    \label{fig:mixinst-main-compare}
    \end{subfigure}
    
    \vspace{-2mm}
    \caption{Regret curves for MixInstruct.}
    \label{fig:mixinst-main}
    \vspace{-4mm}
\end{minipage}
\end{wrapfigure}

To make the regret defined in (\ref{eq:regret}) computable, we reconstruct the utility function $r^{*}(x, a)$.
Pairwise comparisons between LLM candidates are translated into scores by adding a win value of 1, a tie 0.5, and a loss 0.
During the translation process, a Condorcet winner may emerge \citep{black1958,wiki_condorcet_winner}. 
To ensure it receives the highest score, we assign the Condorcet winner a top score with an additional bonus.
During the data analysis stage, we found the existence of ambiguous queries. 
We applied the OpenAI API to assign an ambiguity score to each query and removed the most ambiguous 8\% and 15\% of queries.
The ambiguity removal process introduces additional sub-strings \texttt{\_8} and \texttt{\_15} in the naming of regret curves. 
Using (\ref{eq:def_mix_inst}), the rest of the experiment setup follows the procedure described in \S~\ref{sec:exp_routerb}.
We obtain Fig.~\ref{fig:mixinst-main} and deferred all results to App.~\ref{sec:more_mixinst}.
% due to space limit.

% --- 1x2 layout wrapfigure ---
% --- 1x2 layout wrapfigure ---
% \begin{figure}[h]
%   \centering
%   % Left plot
%   \begin{subfigure}{0.48\textwidth}
%     \centering
%     \includegraphics[width=\linewidth]{plots/mixinst_seed_m1_set1_figure_0.pdf}
%     \caption{Variants of (\ref{eq:def_mix_inst}) versus \texttt{OpenAItext}.}
%     \label{fig:mixinst-main-8}
%   \end{subfigure}\hfill
%   % Right plot
%   \begin{subfigure}{0.48\textwidth}
%     \centering
%     \includegraphics[width=\linewidth]{plots/mixinst_seed_m1_set2_figure_0.pdf}
%     \caption{Ambiguity removal: 8\% versus 15\%.}
%     \label{fig:mixinst-main-compare}
%   \end{subfigure}

%   \caption{MixInstruct: Cumulative regret curves.}
%   \label{fig:mixinst-main}
% \end{figure}

In Fig.~\ref{fig:mixinst-main-8}, we again observe that implementations of the proposed (\ref{eq:def_mix_inst}) outperform \texttt{OpenAItext} variants, demonstrating that our approach remains effective even when the dataset lacks metadata information.
When comparing results under different degrees of ambiguity removal, we observe a consistent pattern in Fig.~\ref{fig:mixinst-main-compare} across \texttt{e5b\_E4}, \texttt{mpnet\_E4}, and \texttt{OpenAItext\_5}: removing the top 15\% of ambiguous queries is worse than removing only the top 8\%. 
This highlights the risk of discarding learnable information when too many queries are removed.

\vspace{-1mm}\section{Concluding Remarks and Future Work}
\label{sec:conclusions_future}

We proposed CCFT, an embedding learning strategy that aligns prospective queries with model expertise through category-calibrated representations. 
Four variants of CCFT were implemented and combined with the theoretically grounded FGTS.CDB algorithm to form the first trainable contextual dueling learner for LLM routing. 
The proposed methods were systematically evaluated on two real-world datasets, RouterBench and MixInstruct, demonstrating their effectiveness. 
Our approach also exhibits robust generalization and achieves a performance-cost balance, both of which are critical for practical deployment.

Looking forward, we highlight two promising directions for future work.
First, as noted in \S~\ref{sec:robust_gen}, our current model representation is effective but may not fully capture the potential of LLM expertise.
Enhancing this alignment between query semantics and model capabilities could lead to even better routing performance. 
We plan to further pursue this direction by exploring new factors and techniques for representing model expertise.
Second, although our method is designed for pairwise feedback, we conjecture that it can be adapted to work with pointwise feedback as well. 
However, building a unified system that can effectively integrate both types of supervision remains an open challenge.
Addressing this would offer both practical value and deeper academic understanding.

% (1) Integrate the preference feedback and rating scale feedback.
% Our method considers dueling feedback currently. 
% It can be easily modified to take rating scale feedback.
% However, a challenge is to integrate both. 
% It is unclear how to build an online learner that takes both forms of input.

% (2) Dynamically retrain the model embedding.
% The proposed model embedding is trained by a fixed offline dataset.
% As time goes on, the service provider will receive new annotated information.
% It would be interesting to construct a mechanism that adapts the model embedding to reflect the newly available dataset.

\paragraph{Ethics statement}
% All authors of this submission have read and agree to abide by the ICLR Code of Ethics. 
This work represents an original contribution derived from our research on LLM routing. We have made an honest and balanced effort to report both the strengths and limitations of the proposed methods. The study is primarily methodological and theoretical in nature. The text embedding models used are either open-source academic resources or publicly available commercial products. All datasets employed are publicly accessible and widely accepted within the research community. Therefore, we do not anticipate any immediate risks of misuse or harm to human society arising from this work.

\paragraph{Reproducibility statement}
We list below the datasets and text embedding models used in the experiments reported in this submission.
\begin{itemize}
    \item MMLU \url{https://openreview.net/forum?id=d7KBjmI3GmQ}
    \item RouterBench \url{https://openreview.net/forum?id=IVXmV8Uxwh}
    \item MixInstruct \url{https://aclanthology.org/2023.acl-long.792/}
    \item text-embedding-3-large of OpenAI \url{https://platform.openai.com/docs/guides/embeddings}
    \item all-MiniLM-L6-v2 of Sentence-Transformers \url{https://huggingface.co/sentence-transformers/all-MiniLM-L6-v2}
    \item all-mpnet-base-v2 of Sentence-Transformers \url{https://huggingface.co/sentence-transformers/all-mpnet-base-v2}
    \item intfloat/e5-base \url{https://arxiv.org/abs/2212.03533}
\end{itemize}
% Upon acceptance, we will release the code necessary to reproduce the experiments, including data processing, embedding learning, and algorithm evaluation. To ensure reproducibility, we have seeded all random processes using the following code block.
We will release the code necessary to reproduce the experiments, including data processing, embedding learning, and algorithm evaluation. To ensure reproducibility, we have seeded all random processes using the following code block.
\begin{listing}[h]
\caption{A Python function to ensure reproducibility.}
\label{code_snippet:set_seed}
\begin{minted}[fontsize=\footnotesize, linenos, breaklines]{python}
def set_seed(seed):
    random.seed(seed)
    np.random.seed(seed)
    torch.manual_seed(seed)
    if torch.cuda.is_available():
        torch.cuda.manual_seed_all(seed)
        torch.backends.cudnn.deterministic = True
\end{minted}
\end{listing}

% Optional.

% \mycmt{All acknowledgments, including those to funding agencies, go at the end of the paper.}

%%%%%%%%%%%%%%%%%%%%%%%%%%%%%%%%%%%%%%%%%%%%%%%%%%%%%%%%%%%%%%%%%%%%%%%%%%%
%%%%%%%%%%%%%%%%%%%%%%%%%%%%%%%%%%%%%%%%%%%%%%%%%%%%%%%%%%%%%%%%%%%%%%%%%%%
\newpage
\bibliography{iclr2026/bibtex_corrected_duel-llm}
% \bibliography{iclr2026/bibtex_duel-llm}
\bibliographystyle{iclr2026_conference}

%%%%%%%%%%%%%%%%%%%%%%%%%%%%%%%%%%%%%%%%%%%%%%%%%%%%%%%%%%%%%%%%%%%%%%%%%%%
%%%%%%%%%%%%%%%%%%%%%%%%%%%%%%%%%%%%%%%%%%%%%%%%%%%%%%%%%%%%%%%%%%%%%%%%%%%
\newpage

\appendix

\vspace{-1mm}\section{Supplementary Materials for Section~\ref{sec:generic_method}}
\label{sec:more_mmlu}

\vspace{-2mm}\subsection{Experimental Details of MMLU}
\label{sec:mmlu_experimental_details}

\begin{figure}[h]
  \centering
  \includegraphics[width=0.48\linewidth]{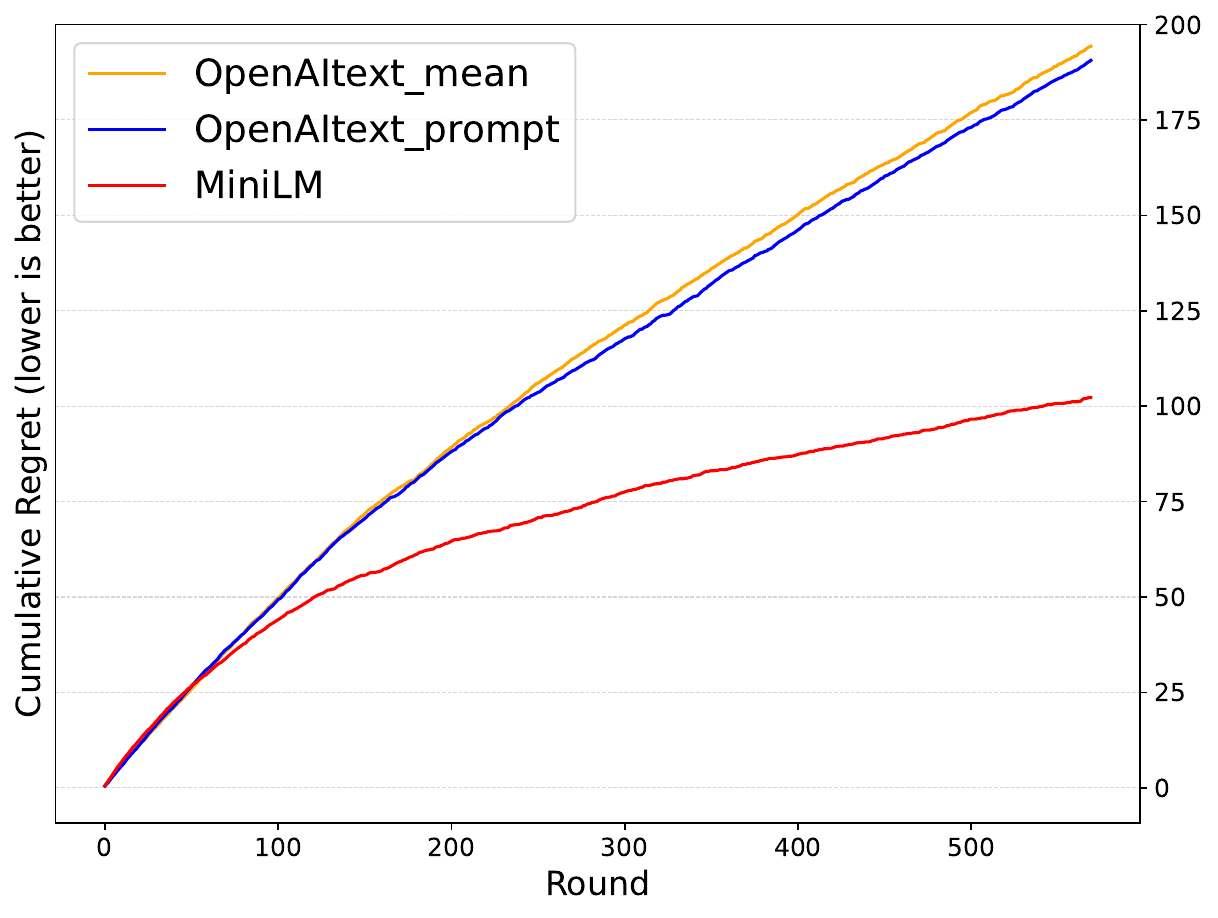}
  \caption{Two failed examples (\texttt{OpenAItext\_mean} and \texttt{OpenAItext\_prompt}) versus a successful example (\texttt{MiniLM}).}
  \label{fig:failure_exs_appendix}
\end{figure}

This section explains how the curves in Fig.~\ref{fig:failure_exs_appendix} are constructed.
We chose five topics, abstract algebra, anatomy, astronomy, international law, and machine learning, from MMLU \citep{hendrycks2021mmlu}.
Queries are sampled to form two disjoint offline learning and online testing sets.
For each topic, ten queries are sampled for offline learning.
The online samples for each topic are drawn in proportion to the dataset, forming an online test set of 595 queries in total.

\paragraph{The Construction of \texttt{OpenAItext\_prompt} and \texttt{OpenAItext\_mean}}
Since MMLU does not involve LLMs, we need to construct our own LLM experts and the corresponding performance values.
A straightforward way is to assume there are five LLMs, each with expertise in one of the topics.
Using offline queries, we explore two approaches to generate the model embeddings $a_k$\footnote{For consistency, we use $a_k$ both to index an LLM (as in \S~\ref{sec:background}) and to denote its embedding (as explained next), with the intended meaning clear from context.}.
In the first way, we encode the model description via OpenAI's \texttt{text-embedding-3-large}, where the model description is the combination of our handcrafted prompt with offline queries (Listing~\ref{code_snippet:prompt_mmlu} in App.~\ref{sec:prompts}).
We use \texttt{OpenAItext\_prompt} to denote its results.
In the second way, we first generate offline query embeddings of a topic using \texttt{text-embedding-3-large} and then take their average as the model embedding. 
\texttt{OpenAItext\_mean} is used to denote its results.
The intuition behind both approaches is to represent a model using the sample queries it excels at.
It is rather simple to generate a query embedding $x$, as we simply feed the query string into \texttt{text-embedding-3-large}.
We tested the Hadamard product (element-wise multiplication) $\phi(x, a) = x * a$ with normalization and vector addition (element-wise addition) $\phi(x, a) = x + a$ with normalization to construct $\phi$ and keep the first one based on the experimental outcomes.
To construct performance values, we compute a similarity matrix using the average query embeddings for each topic and the cosine similarity function. 
Given the topic of the current query and the algorithm’s selections, we can retrieve the corresponding similarity scores from this matrix to quantify the algorithm’s performance. 
These similarity scores are then used both to sample feedback via the BTL model and to generate the performance values needed for computing regret at each round.

\paragraph{The Construction of \texttt{MiniLM}}
\texttt{MiniLM} is an abbreviation of \texttt{all-MiniLM-L6-v2}.
Its construction follows the procedure as \texttt{OpenAItext\_mean}, with two modifications.
First, the embedding model is replaced with \texttt{all-MiniLM-L6-v2}.
Second, contrastive learning \citep{Khosla20SCL, Nils2019sentenceBERT} is applied to fine-tune the model. 
To do so, we construct similar and dissimilar query pairs based on their source category, and fine-tune the model using a cosine similarity loss for four epochs.
The regret curve corresponding to this model is labeled as \texttt{MiniLM}.

We note that Fig.~\ref{fig:failure_exs} is identical to Fig.~\ref{fig:failure_exs_appendix}, and that \texttt{MiniLM} is omitted from the main text.
The goal there is to illustrate how a successful routing method should behave, rather than to define the method itself. 
Since \texttt{MiniLM} is not the final version of our proposed approach, and MMLU is not an ideal benchmark for evaluating routing strategies, we chose to defer these details to the appendix. 
Nonetheless, MMLU is sufficiently simple to serve as a synthetic simulation for demonstrating our motivation.

\vspace{-2mm}\subsection{Proof of Proposition~\ref{thm:weight_without_scores}}
\label{proof:weight_without_scores}

\begin{proof}[Proof of Proposition~\ref{thm:weight_without_scores}]
    Let $k$ be fixed.
    Assume without loss of generality that each $f_{km} n$ is an integer, the size of $\mathcal{G}_k$ is $\sum_{j=1}^{M} f_{kj} n$.
    Then,
    \eqary{
        \frac{\sum_{q \in \mathcal{G}_k} q}{|\mathcal{G}_k|}
        =\ &
        \frac{\sum_{m=1}^{M} \sum_{q \in \mathcal{G}_k\cap\mathcal{Q}_m} q}{\sum_{j=1}^{m} f_{kj} n} 
        =\ 
        \sum_{m=1}^{M} \frac{f_{km} n}{\sum_{j=1}^{M} f_{kj} n} \left( \frac{\sum_{q \in \mathcal{G}_k\cap\mathcal{Q}_m} q}{f_{km} n} \right).
    }
    The term in the parentheses is the sample average of $f_{km} n$ independent embeddings drawn from category $m$. 
    Hence, it is an unbiased estimator of $\mathbb{E}[Q_m]$ and we have the proposition.
    % Hence, we have the proposition by noting that $\frac{\sum_{q \in \mathcal{G}_k\cap\mathcal{Q}_m} q}{f_{km} n}$ is an unbiased estimation for $\mathbb{E}[Q_m]$.
\end{proof}

\vspace{-2mm}\subsection{t-SNE Visualizations}

\begin{figure*}[h]
  \centering
  \includegraphics[width=0.95\linewidth]{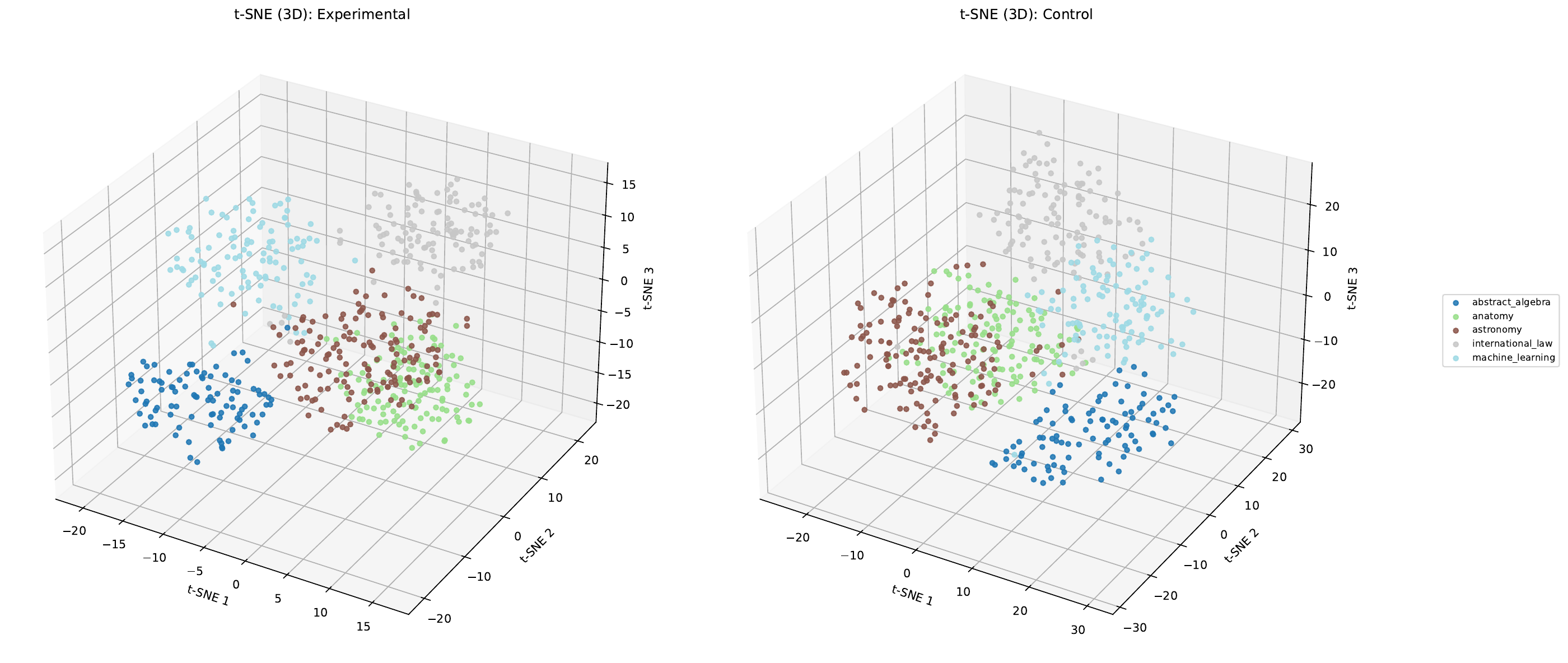}
  \caption{3D t-SNE visualization of the embeddings generated by the fine-tuned MiniLM model discussed in Appendix~\ref{sec:mmlu_experimental_details} (left), compared to those without fine-tuning (right). Each point represents an embedding projected into 3D space, with colors indicating cluster membership.}
  \label{fig:tsne-3d}
\end{figure*}
% \mycmt{t-SNE and 3D t-SNE plots?}

\vspace{-1mm}\section{RouterBench Supplementary Materials}
\label{sec:more_routerb}

\vspace{-2mm}\subsection{Table 1 of \cite{hu2024routerbench}}
\label{sec:more_categorical_weighting}

Tab.~\ref{table:Table_1_routerb} is identical to Table 1 in \cite{hu2024routerbench}. 
We include it here for completeness.

\begin{table}[htbp]
\centering
\caption{Performance and cost across benchmarks}
\label{table:Table_1_routerb}
\begin{adjustbox}{max width=\textwidth}
\begin{tabular}{lcccccccccccccc}
\toprule
\textbf{LLM} & 
\multicolumn{2}{c}{\textbf{MMLU}} & 
\multicolumn{2}{c}{\textbf{MT-Bench}} & 
\multicolumn{2}{c}{\textbf{MBPP}} & 
\multicolumn{2}{c}{\textbf{HellaSwag}} & 
\multicolumn{2}{c}{\textbf{Winogrande}} & 
\multicolumn{2}{c}{\textbf{GSM8k}} & 
\multicolumn{2}{c}{\textbf{ARC}} \\
\cmidrule(lr){2-3}
\cmidrule(lr){4-5}
\cmidrule(lr){6-7}
\cmidrule(lr){8-9}
\cmidrule(lr){10-11}
\cmidrule(lr){12-13}
\cmidrule(lr){14-15}
& Perf$\uparrow$ & Cost$\downarrow$ 
& Perf$\uparrow$ & Cost$\downarrow$ 
& Perf$\uparrow$ & Cost$\downarrow$ 
& Perf$\uparrow$ & Cost$\downarrow$ 
& Perf$\uparrow$ & Cost$\downarrow$ 
& Perf$\uparrow$ & Cost$\downarrow$ 
& Perf$\uparrow$ & Cost$\downarrow$ \\
\midrule
WizardLM 13B     & 0.568 & 0.122 & 0.796 & 0.006 & 0.364 & 0.011 & 0.636 & 0.727 & 0.512 & 0.040 & 0.510 & 0.354 & 0.660 & 0.068 \\
Mistral 7B       & 0.562 & 0.081 & 0.779 & 0.003 & 0.349 & 0.006 & 0.541 & 0.485 & 0.562 & 0.027 & 0.409 & 0.210 & 0.642 & 0.046 \\
Mixtral 8x7B     & 0.733 & 0.245 & 0.921 & 0.012 & 0.573 & 0.023 & 0.707 & 1.455 & 0.677 & 0.081 & 0.515 & 0.594 & 0.844 & 0.137 \\
Code Llama 34B   & 0.569 & 0.317 & 0.796 & 0.015 & 0.465 & 0.021 & 0.525 & 1.882 & 0.617 & 0.104 & 0.462 & 0.752 & 0.644 & 0.177 \\
Yi 34B           & 0.743 & 0.326 & 0.938 & 0.018 & 0.333 & 0.031 & 0.931 & 1.938 & 0.748 & 0.107 & 0.552 & 0.867 & 0.882 & 0.182 \\
GPT-3.5          & 0.720 & 0.408 & 0.908 & 0.026 & 0.651 & 0.044 & 0.816 & 2.426 & 0.630 & 0.134 & 0.601 & 1.170 & 0.855 & 0.228 \\
Claude Instant V1& 0.384 & 0.327 & 0.863 & 0.030 & 0.550 & 0.064 & 0.801 & 1.943 & 0.512 & 0.108 & 0.626 & 1.300 & 0.821 & 0.183 \\
Llama 70B        & 0.647 & 0.367 & 0.854 & 0.022 & 0.302 & 0.039 & 0.736 & 2.183 & 0.504 & 0.121 & 0.529 & 0.870 & 0.794 & 0.205 \\
Claude V1        & 0.475 & 3.269 & 0.938 & 0.361 & 0.527 & 0.607 & 0.841 & 19.43 & 0.570 & 1.077 & 0.653 & 11.09 & 0.889 & 1.829 \\
Claude V2        & 0.619 & 3.270 & 0.854 & 0.277 & 0.605 & 0.770 & 0.421 & 19.50 & 0.446 & 1.081 & 0.664 & 13.49 & 0.546 & 1.833 \\
GPT-4            & 0.828 & 4.086 & 0.971 & 0.721 & 0.682 & 1.235 & 0.923 & 24.29 & 0.858 & 1.346 & 0.654 & 19.08 & 0.921 & 2.286 \\
\bottomrule
\end{tabular}
\end{adjustbox}
\end{table}

% Tables \ref{table:Table_1_routerb} and \ref{table:weighting} are from \url{https://chatgpt.com/c/68cb776a-3fbc-8329-b5a1-3cf59d40b1f5}

% \begin{table}[h!]
% \label{table:hypothetic_method}
% \centering
% \caption{LLM Scores Across Categories}
% \begin{tabular}{lccccccc}
% \toprule
%  & Category 1 & Category 2 & Category 3 & Category 4 & Category 5 & Category 6 & Category 7 \\
%   & $\xi_1$ & $\xi_2$ & $\xi_3$ & $\xi_4$ & $\xi_5$ & $\xi_6$ & $\xi_7$ \\
% \midrule
% LLM $a_1$ & 0.368 & 0.862 & 0.547 & \textbf{0.704} & 0.507 & \textbf{0.561} & 0.812 \\
% LLM $a_2$ & 0.629 & 0.853 & 0.300 & 0.627 & 0.498 & 0.486 & 0.784 \\
% LLM $a_3$ & \textbf{0.721} & \textbf{0.920} & \textbf{0.572} & 0.634 & \textbf{0.673} & 0.485 & \textbf{0.837} \\
% LLM $a_4$ & \textbf{0.727} & \textbf{0.937} & 0.331 & \textbf{0.834} & \textbf{0.743} & \textbf{0.509} & \textbf{0.873} \\
% LLM $a_5$ & 0.456 & 0.840 & \textbf{0.567} & -0.554 & 0.392 & -0.011 & 0.454 \\
% \bottomrule
% \end{tabular}
% \end{table}

% Using the hypothetical table above, we present 
% \bcmt{(maybe move to appendix or delete).}

\vspace{-2mm}\subsection{Additional Results for RouterBench}
\label{sec:more_routerb_results}

\begin{figure*}[h]
  \centering
  % Row 1
  \begin{subfigure}{0.48\textwidth}
    \centering
    \includegraphics[width=\linewidth]{plots/routerb_seed1_m42_set1_figure_0.pdf}
    \caption{\texttt{e5b\_E4} results.}
    \label{fig:grid-rtb-models-all-a}
  \end{subfigure}
  \hfill
  \begin{subfigure}{0.48\textwidth}
    \centering
    \includegraphics[width=\linewidth]{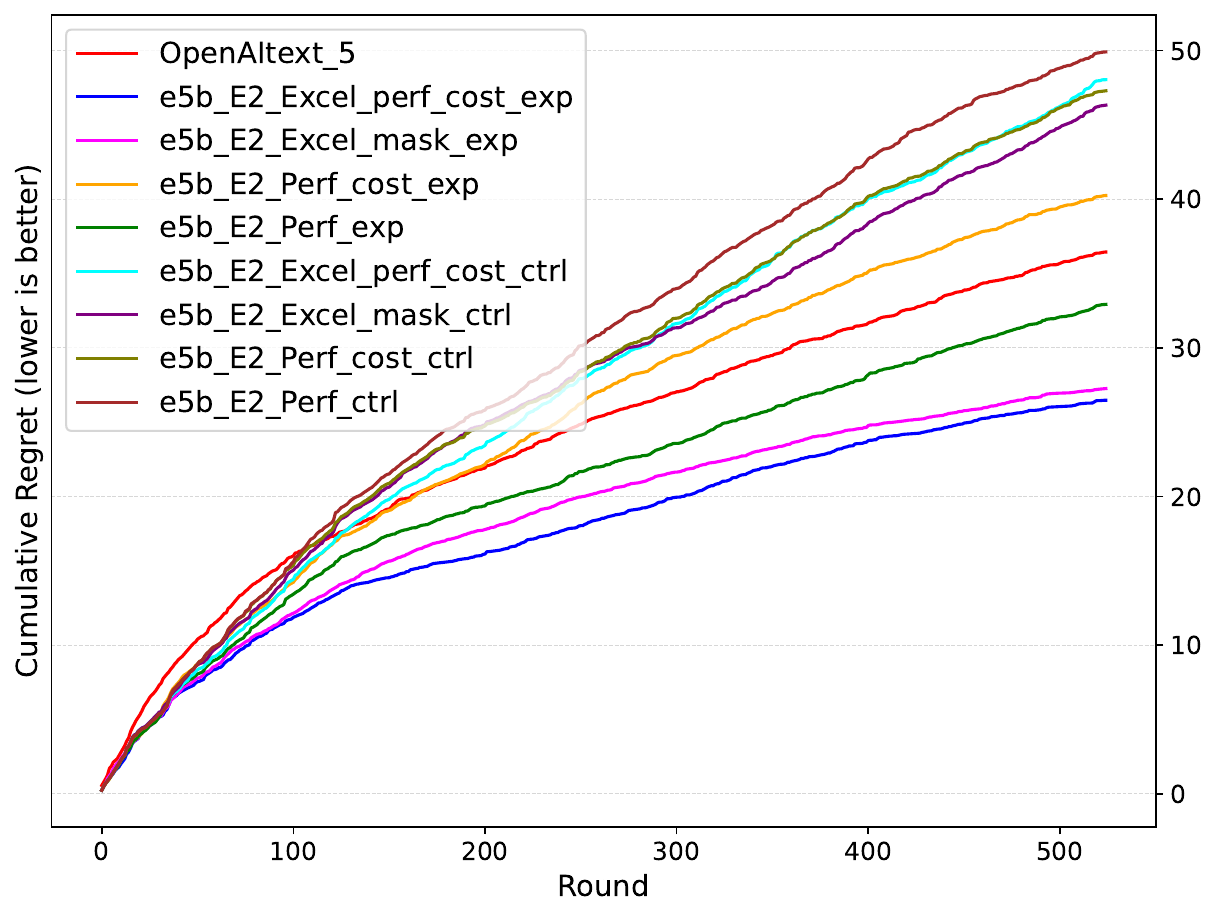}
    \caption{\texttt{e5b\_E2} results.}
    \label{fig:grid-rtb-models-all-b}
  \end{subfigure}

  \vspace{0.3em}

  % Row 2
  \begin{subfigure}{0.48\textwidth}
    \centering
    \includegraphics[width=\linewidth]{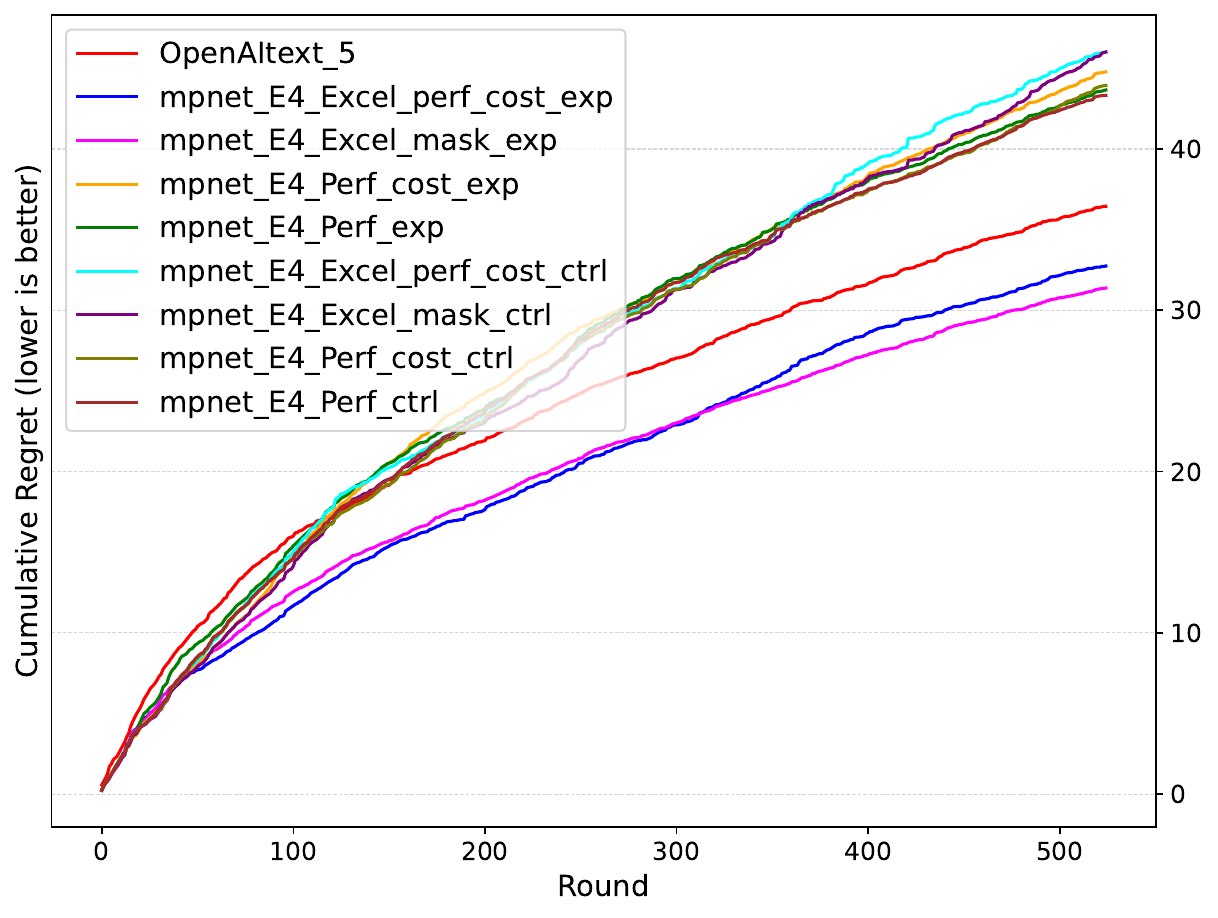}
    \caption{\texttt{mpnet\_E4} results.}
  \end{subfigure}
  \hfill
  \begin{subfigure}{0.48\textwidth}
    \centering
    \includegraphics[width=\linewidth]{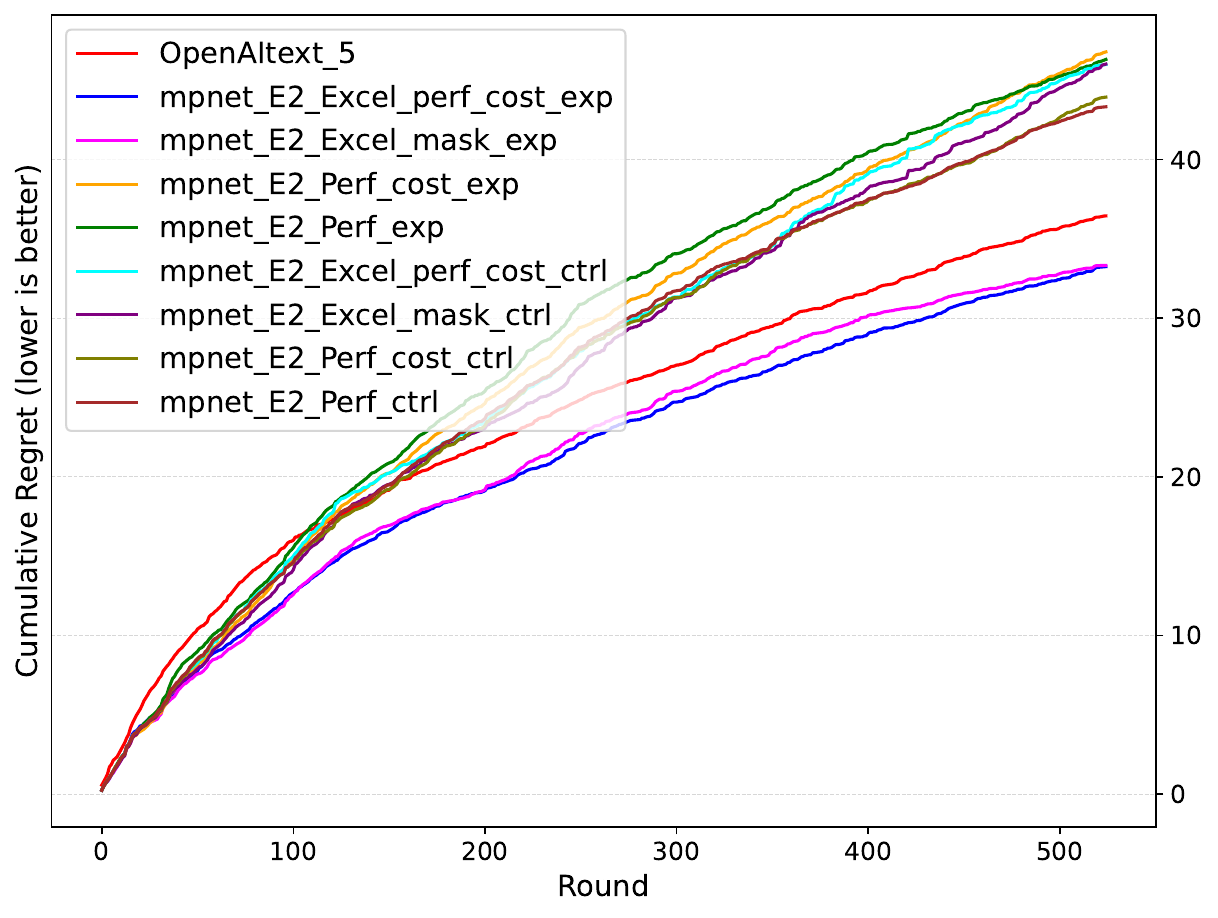}
    \caption{\texttt{mpnet\_E2} results.}
  \end{subfigure}

  \vspace{0.3em}

  % Row 3
  \begin{subfigure}{0.48\textwidth}
    \centering
    \includegraphics[width=\linewidth]{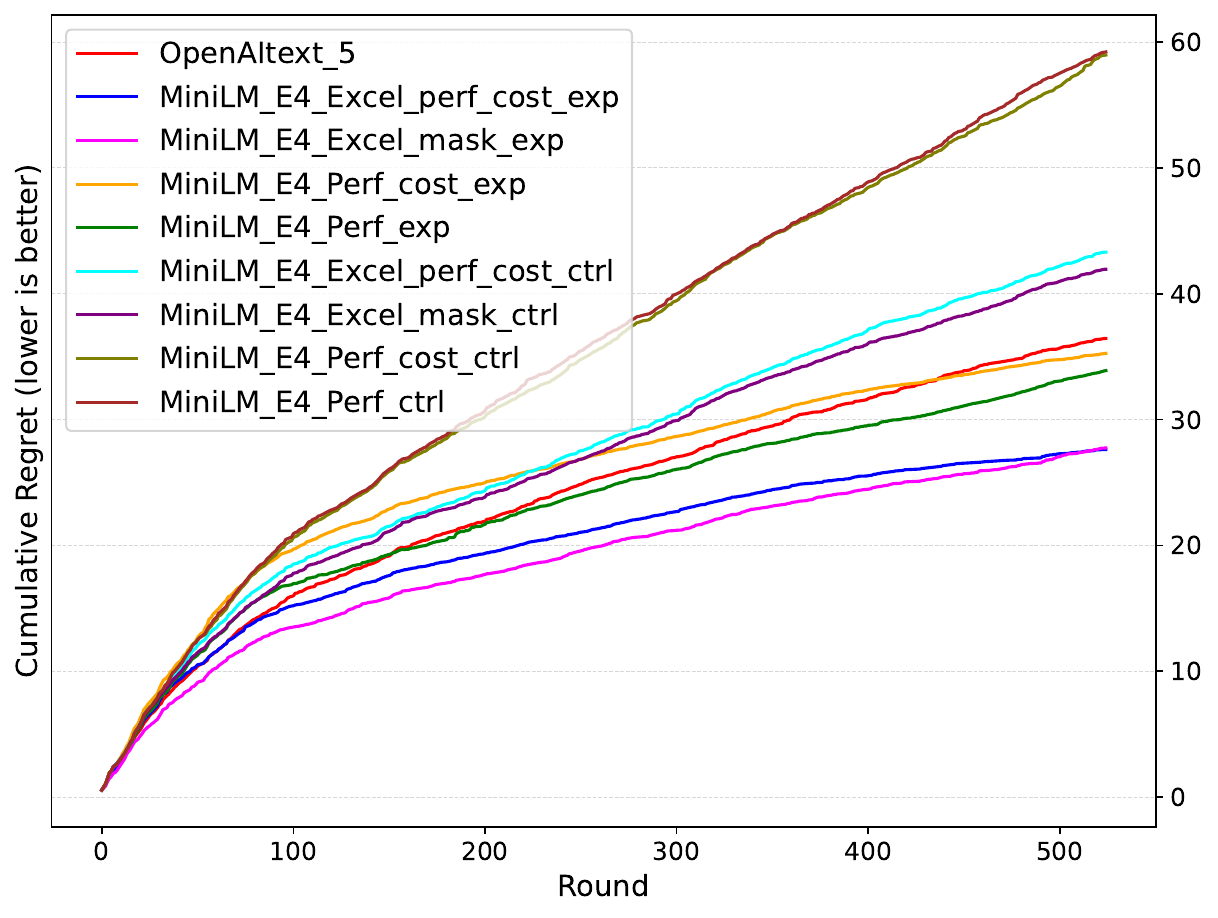}
    \caption{\texttt{MiniLM\_E4} results.}
  \end{subfigure}
  \hfill
  \begin{subfigure}{0.48\textwidth}
    \centering
    \includegraphics[width=\linewidth]{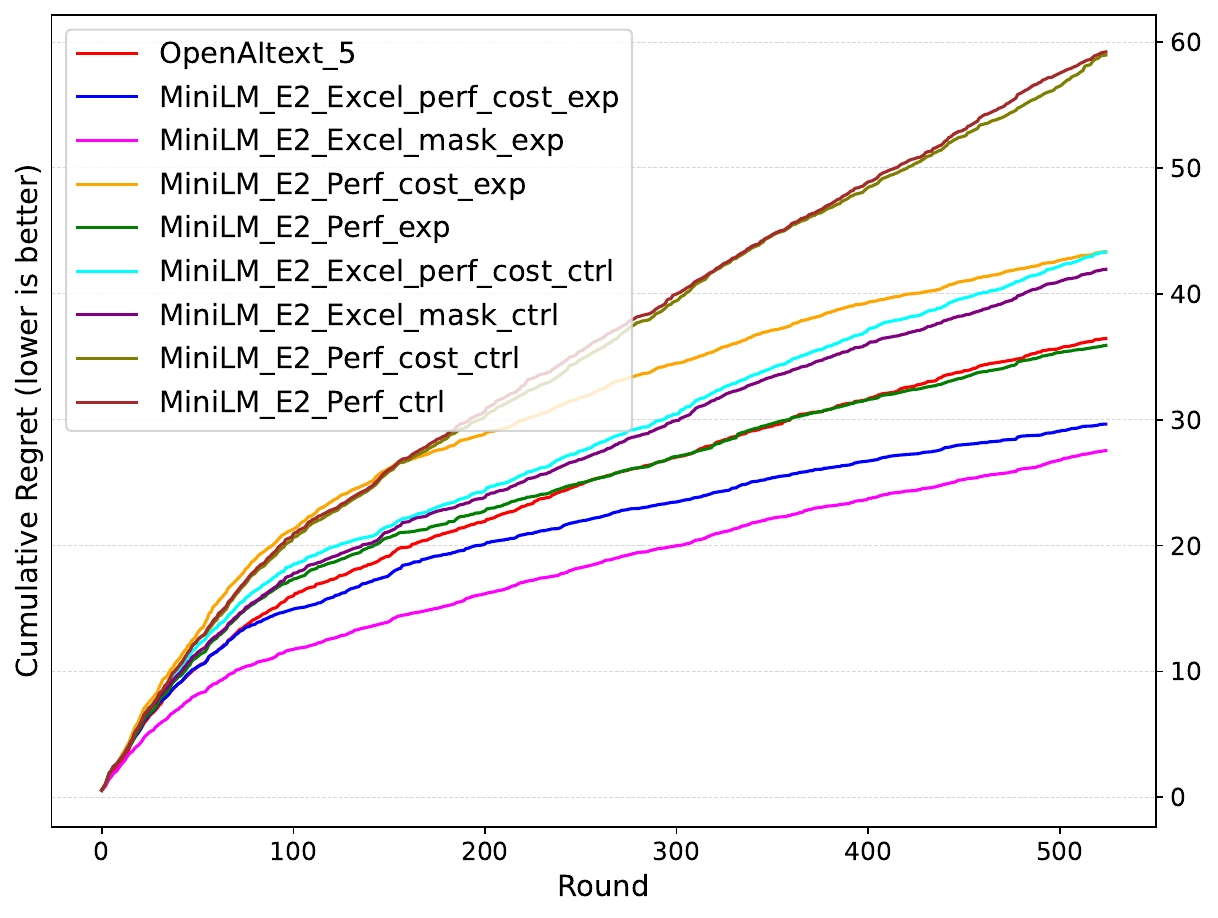}
    \caption{\texttt{MiniLM\_E2} results.}
    \label{fig:grid-rtb-models-all-f}
  \end{subfigure}
  
  \vspace{0.3em}
  
  % Row 4
  \begin{subfigure}{0.48\textwidth}
    \centering
    \includegraphics[width=\linewidth]{plots/routerb_seed2_m42_set4_figure_0.pdf}
    \caption{\texttt{OpenAItext} results.}
    \label{fig:grid-rtb-models-all-g}
  \end{subfigure}
  \hfill
  \begin{subfigure}{0.48\textwidth}
    \centering
    \includegraphics[width=\linewidth]{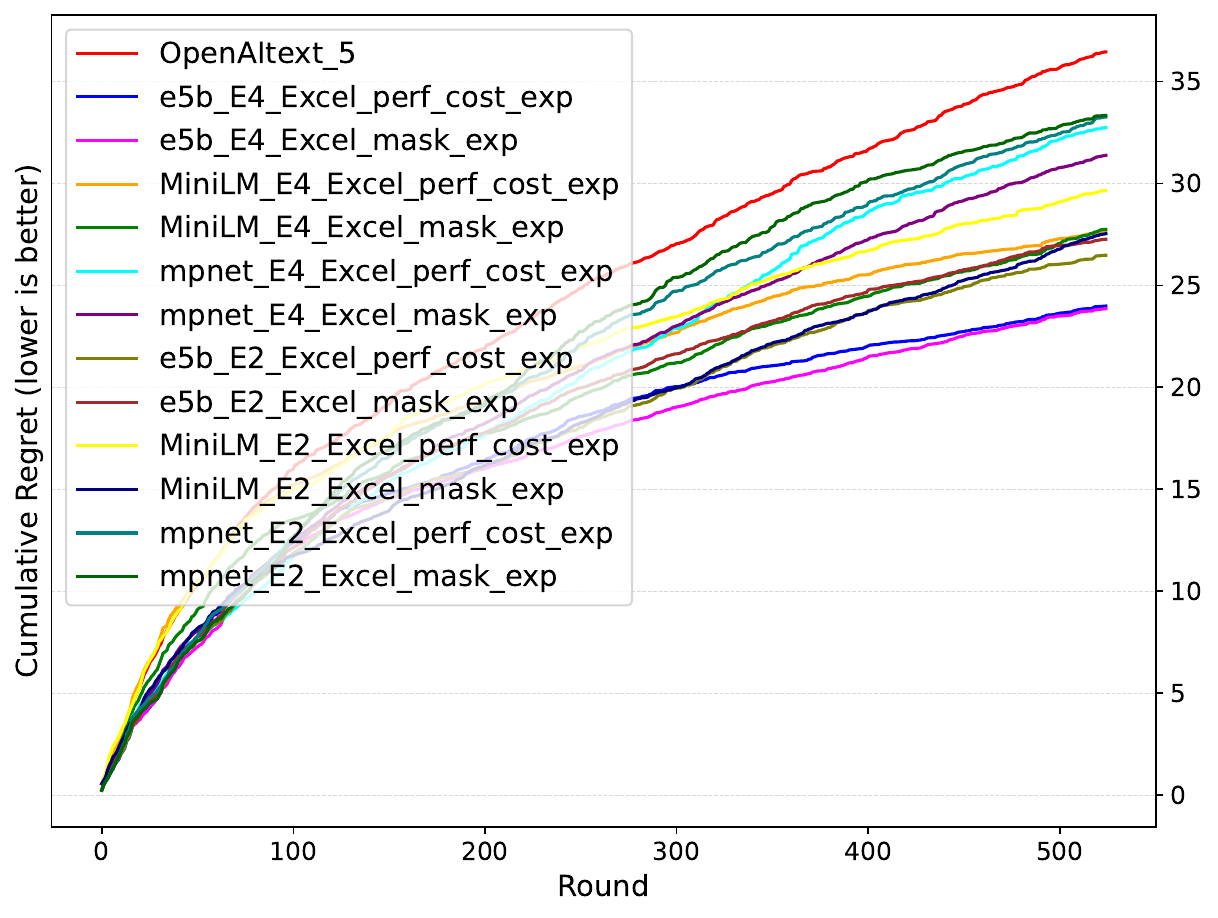}
    \caption{Compare against all text models.}
    \label{fig:grid-rtb-models-all-h}
  \end{subfigure}

  \caption{Cumulative regret curves for RouterBench.}
  \label{fig:grid-rtb-models-all}
\end{figure*}

Fig.~\ref{fig:grid-rtb-models-all} presents the cumulative regret curves for \texttt{e5b\_E4}, \texttt{e5b\_E2}, \texttt{mpnet\_E4}, \texttt{mpnet\_E2}, \texttt{MiniLM\_E4}, \texttt{MiniLM\_E2}, and the \texttt{OpenAItext} variants. Fig.~\ref{fig:grid-rtb-models-all-h} compares all embedding models under the \texttt{Excel\_perf\_cost} and \texttt{Excel\_mask} mechanisms, which represent the most effective weighting methods in most cases.
Note that Fig.~\ref{fig:grid-rtb-models-all-a} is identical to Fig.~\ref{fig:rtb-main-b}, and Fig.~\ref{fig:grid-rtb-models-all-g} is identical to Fig.~\ref{fig:rtb-main-a}.
They are included here for completeness.

% --- 3-by-2 Figure ---
% --- 3-by-2 Figure ---

% % --- 2-by-3 Figure ---
% \begin{figure*}[t]
%   \centering
%   \begin{subfigure}{0.32\textwidth}
%     \centering
%     \includegraphics[width=\linewidth]{plots/routerb_seed1_m42_set1_figure_0.pdf}
%     \caption{A}
%   \end{subfigure}\hfill
%   \begin{subfigure}{0.32\textwidth}
%     \centering
%     \includegraphics[width=\linewidth]{plots/routerb_seed1_m42_set1_figure_1.pdf}
%     \caption{B}
%   \end{subfigure}\hfill
%   \begin{subfigure}{0.32\textwidth}
%     \centering
%     \includegraphics[width=\linewidth]{plots/routerb_seed1_m42_set1_figure_2.pdf}
%     \caption{C}
%   \end{subfigure}

%   \vspace{0.3em}

%   \begin{subfigure}{0.32\textwidth}
%     \centering
%     \includegraphics[width=\linewidth]{plots/routerb_seed1_m42_set1_figure_3.pdf}
%     \caption{D}
%   \end{subfigure}\hfill
%   \begin{subfigure}{0.32\textwidth}
%     \centering
%     \includegraphics[width=\linewidth]{plots/routerb_seed1_m42_set1_figure_4.pdf}
%     \caption{E}
%   \end{subfigure}\hfill
%   \begin{subfigure}{0.32\textwidth}
%     \centering
%     \includegraphics[width=\linewidth]{plots/routerb_seed1_m42_set1_figure_5.pdf}
%     \caption{F}
%   \end{subfigure}

%   \caption{Cumulative regret curves.}
%   \label{fig:grid-rtb-models}
% \end{figure*}
% % --- 2-by-3 Figure ---

\vspace{-2mm}\subsection{Comparison with MixLLM}
\label{sec:more_routerb_compare}

% \mycmt{(1) Compare our TS vs MixLLM's UCB approach \citep{wang2025mixllm}; (2) Compare with Pareto optimization approaches.}
% \new
% Before ending the section, we compare our work with the most relevant MixLLM of \cite{wang2025mixllm}.
We compare our work with the most relevant related method, MixLLM, proposed by \cite{wang2025mixllm}.
Both approaches adopt online learning frameworks with binary feedback. 
However, there are three fundamental differences between the two.

First, MixLLM uses pointwise feedback (e.g., like/dislike) as input, whereas our method relies on pairwise feedback (i.e., preference comparisons). As a result, the problem settings are inherently different.
Second, MixLLM employs an upper confidence bound (UCB)-based strategy, where uncertainty is managed via the matrix $A_l$ (see (9) in their paper). 
In contrast, our approach is based on TS, where uncertainty is governed by posterior sampling and the likelihood function $L^{j}$ (\ref{eq:likelihood}).
Third, our method requires significantly fewer offline training samples. Specifically, we use only five queries per benchmark (thirty-five in total) for offline learning. 
According to Table 1 in \cite{wang2025mixllm}, MixLLM requires at least 30\% of the dataset for offline training.
This sample efficiency is an appealing feature of our approach.\footnote{In particular, we use five queries per category in the current section, fifteen for robust generalization evaluation (\S~\ref{sec:robust_gen}), and ten queries for the MixInstruct experiments (\S~\ref{sec:exp_mixInst}).}

\vspace{-2mm}\subsection{Additional Results for Robust Generalization}
\label{sec:more_robust_results}

\begin{figure*}[h]
  \centering
  % Row 1
  \begin{subfigure}{0.48\textwidth}
    \centering
    \includegraphics[width=\linewidth]{plots/routerb_rbst1_m42_set1_figure_0.pdf}
    \caption{\texttt{e5b\_E4} results.}
    \label{fig:rtb-rbst-all-a}
  \end{subfigure}
  \hfill
  \begin{subfigure}{0.48\textwidth}
    \centering
    \includegraphics[width=\linewidth]{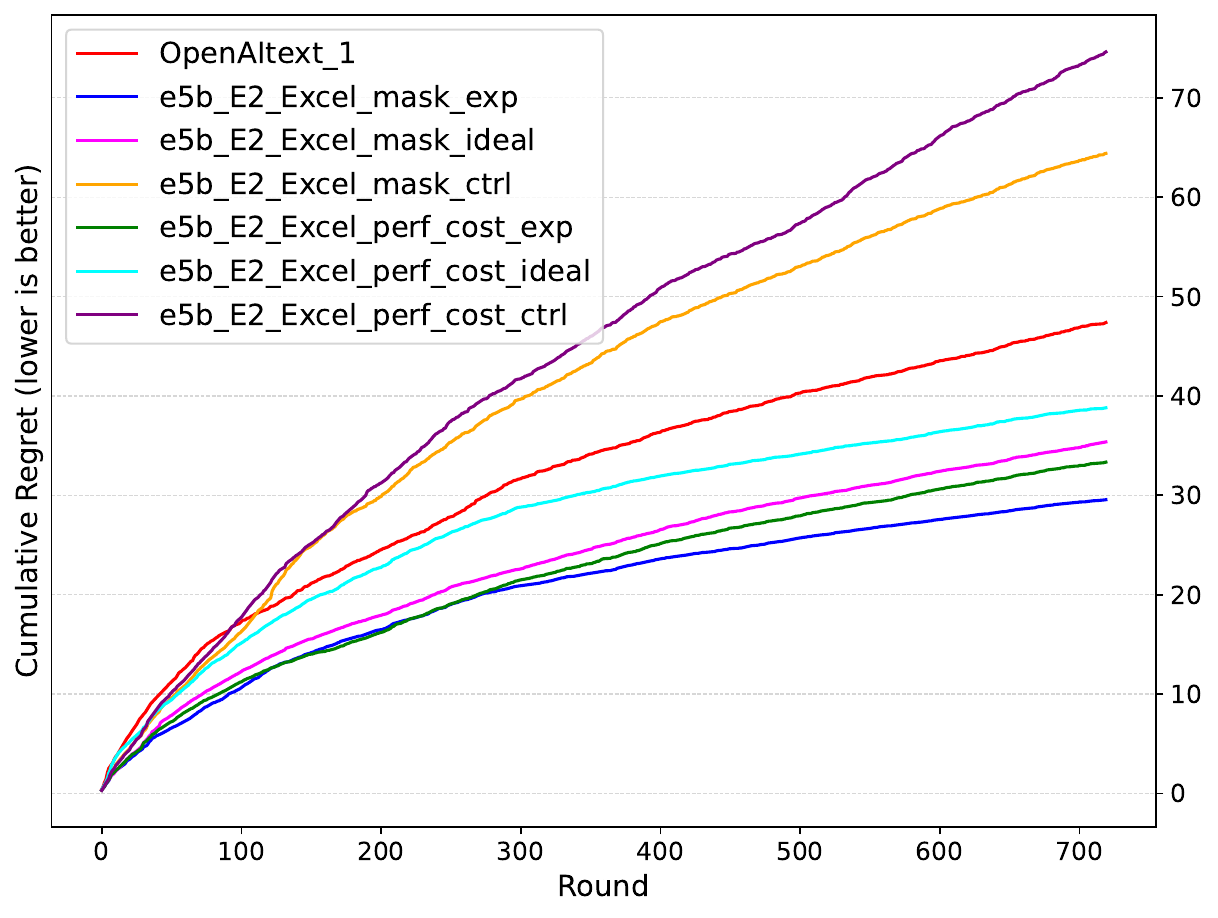}
    \caption{\texttt{e5b\_E2} results.}
    \label{fig:rtb-rbst-all-b}
  \end{subfigure}

  \vspace{0.3em}

  % Row 2
  \begin{subfigure}{0.48\textwidth}
    \centering
    \includegraphics[width=\linewidth]{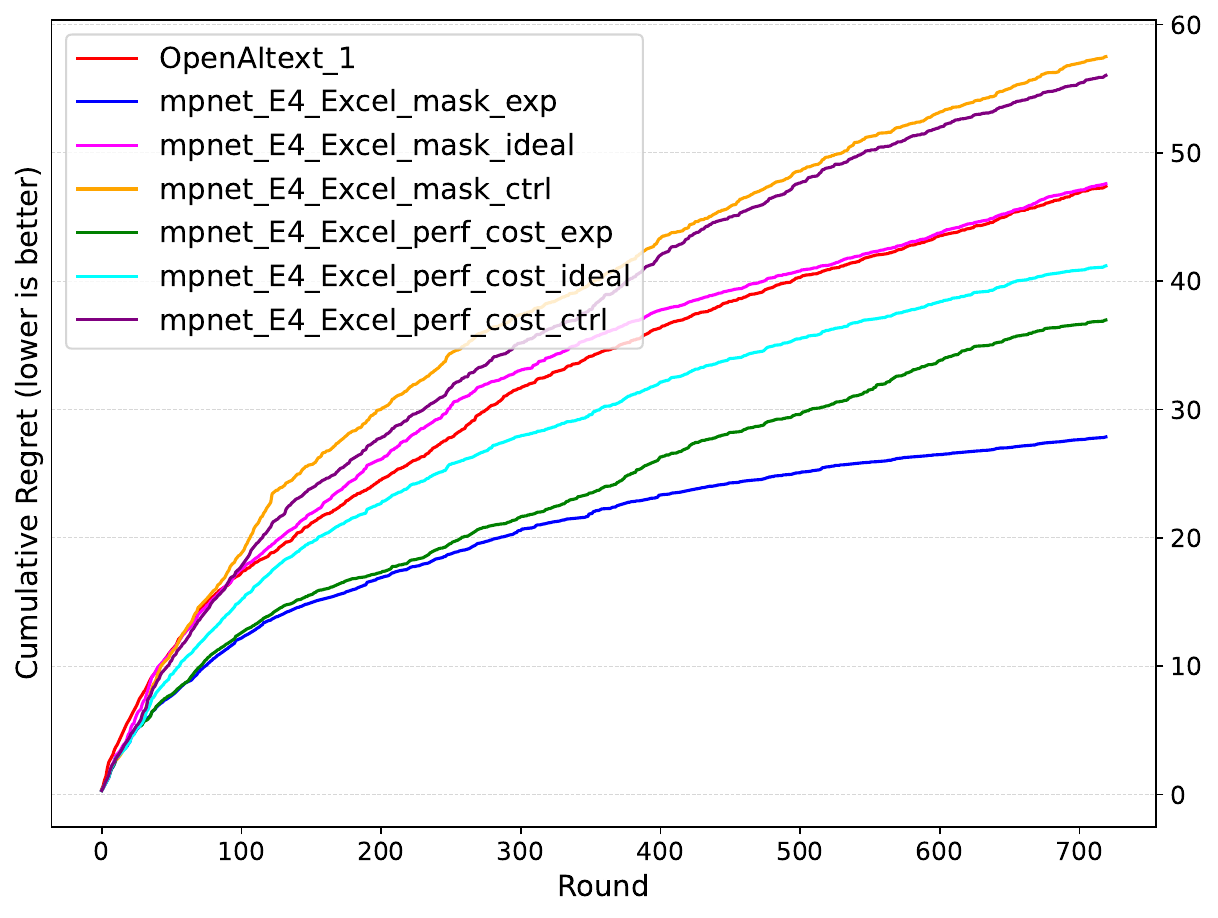}
    \caption{\texttt{mpnet\_E4} results.}
    \label{fig:rtb-rbst-all-c}
  \end{subfigure}
  \hfill
  \begin{subfigure}{0.48\textwidth}
    \centering
    \includegraphics[width=\linewidth]{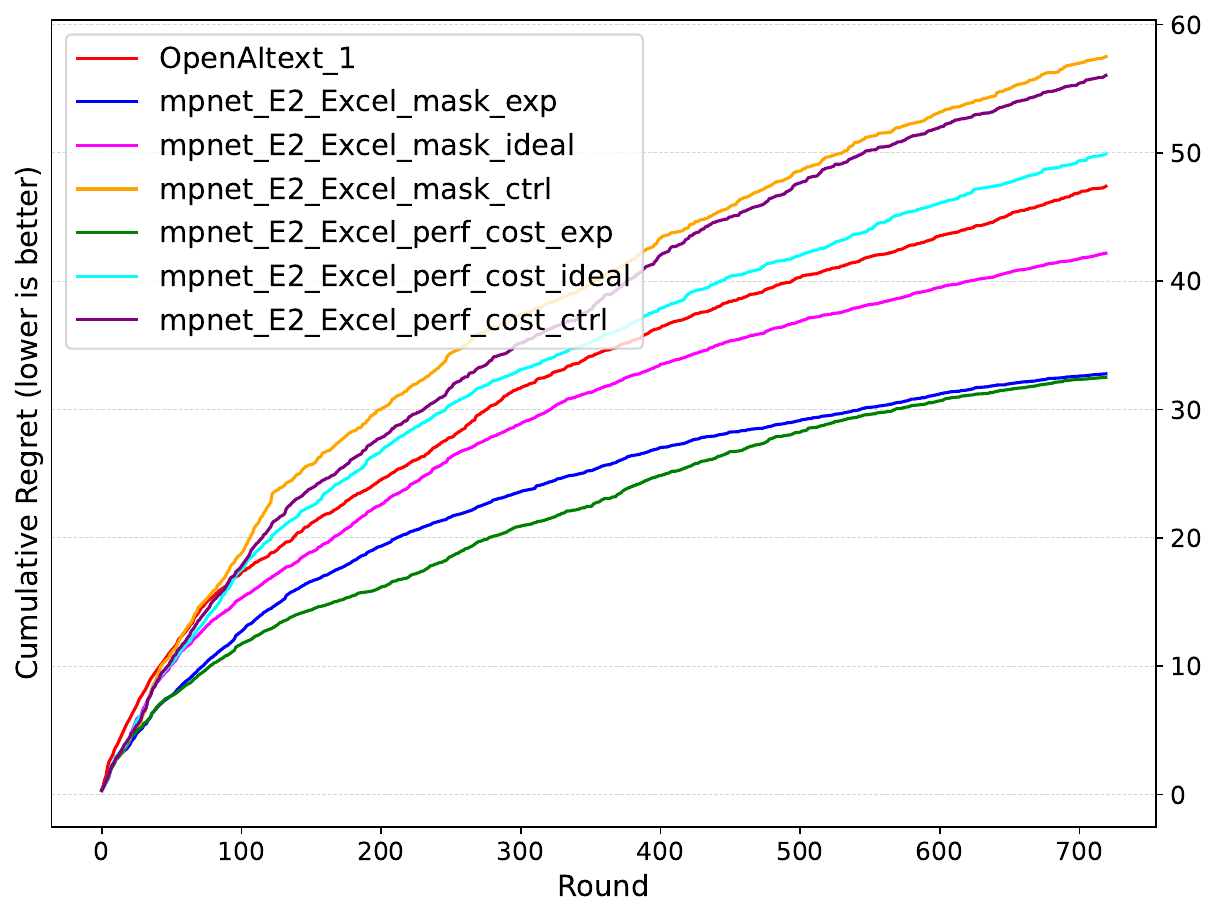}
    \caption{\texttt{mpnet\_E2} results.}
    \label{fig:rtb-rbst-all-d}
  \end{subfigure}

  \vspace{0.3em}

  % Row 3
  \begin{subfigure}{0.48\textwidth}
    \centering
    \includegraphics[width=\linewidth]{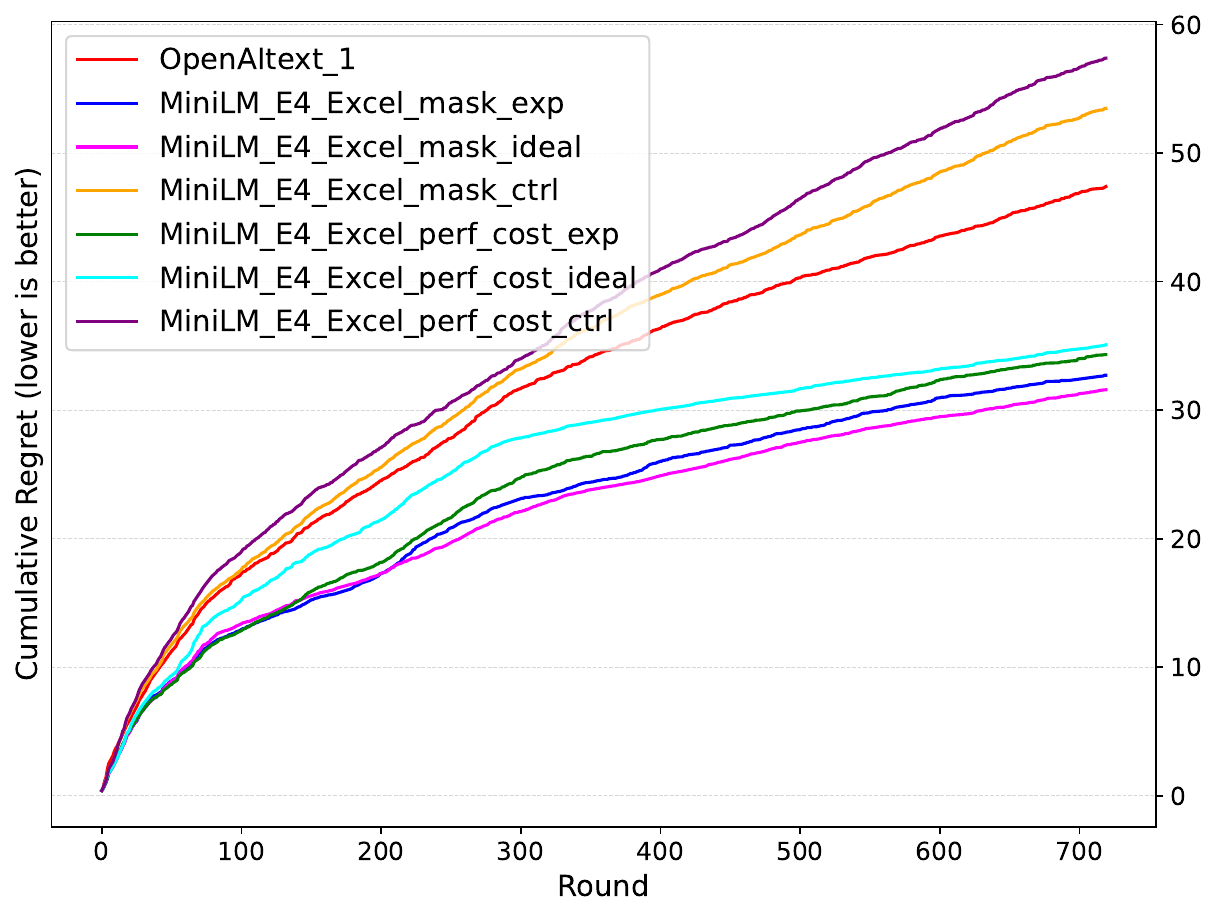}
    \caption{\texttt{MiniLM\_E4} results.}
    \label{fig:rtb-rbst-all-e}
  \end{subfigure}
  \hfill
  \begin{subfigure}{0.48\textwidth}
    \centering
    \includegraphics[width=\linewidth]{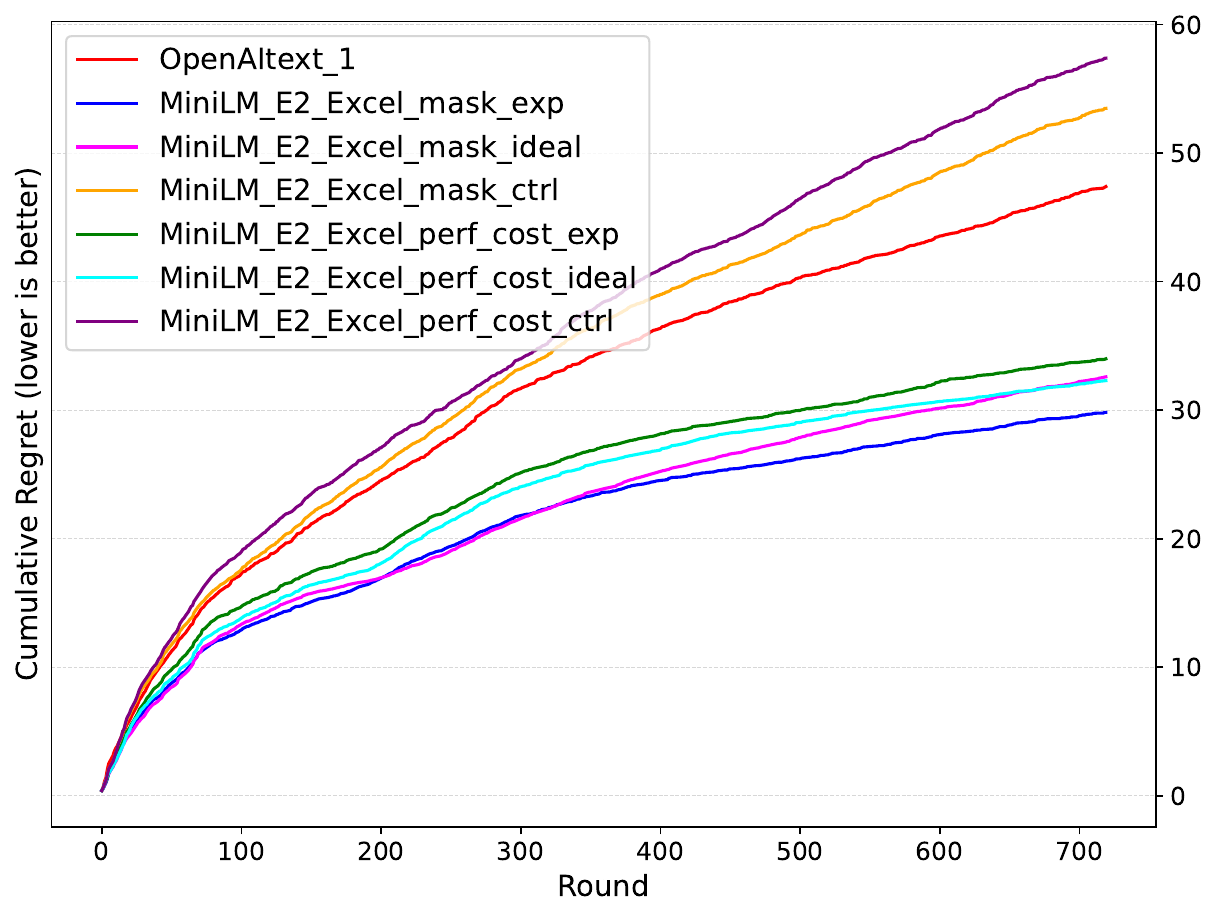}
    \caption{\texttt{MiniLM\_E2} results.}
    \label{fig:rtb-rbst-all-f}
  \end{subfigure}

  \vspace{0.3em}

  % Row 4
  \begin{subfigure}{0.48\textwidth}
    \centering
    \includegraphics[width=\linewidth]{plots/routerb_rbst2_m42_set4_figure_0.pdf}
    \caption{\texttt{OpenAItext} results.}
    \label{fig:rtb-rbst-all-g}
  \end{subfigure}
  \hfill
  \begin{subfigure}{0.48\textwidth}
    \centering
    \includegraphics[width=\linewidth]{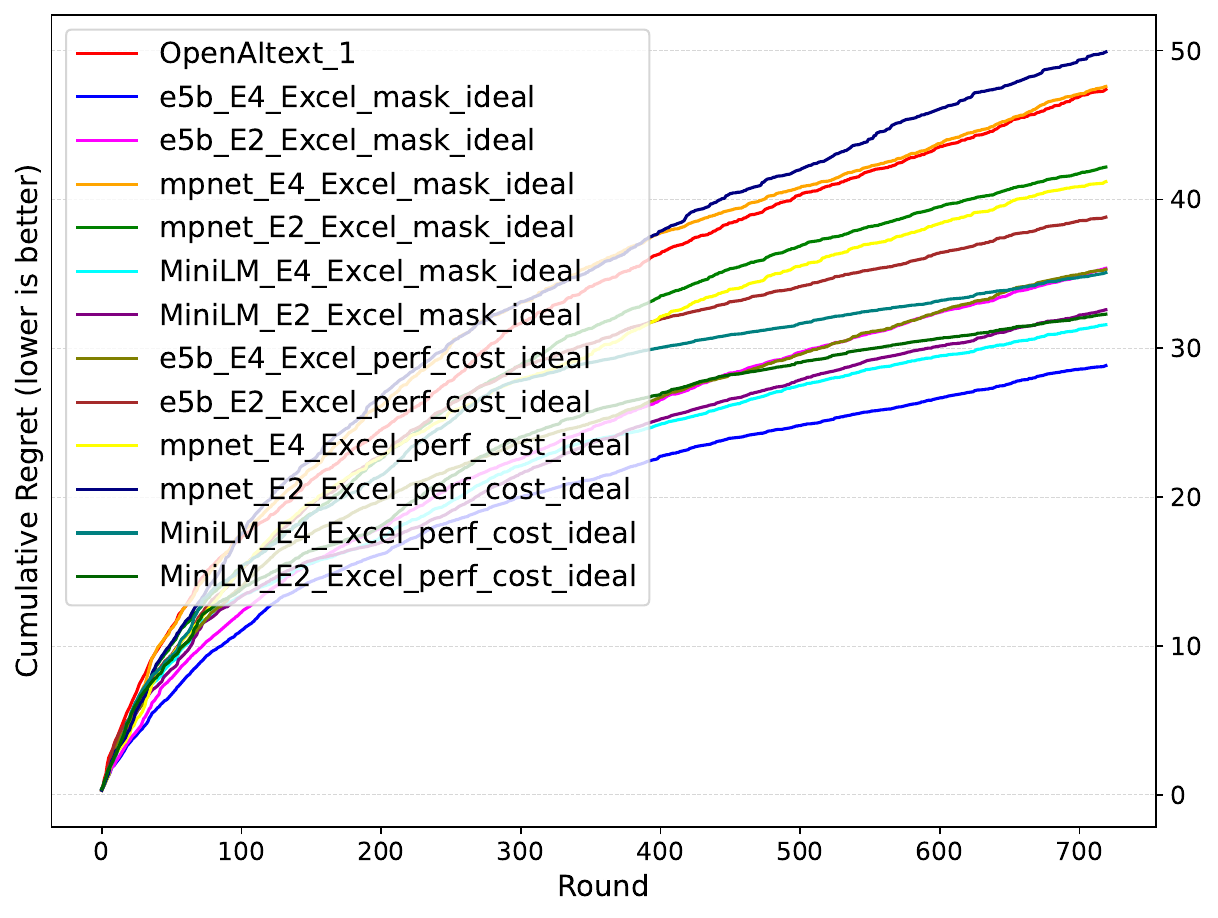}
    \caption{Compare against all text models.}
    \label{fig:rtb-rbst-all-h}
  \end{subfigure}

  \caption{Cumulative regret curves for robust generalization.}
  \label{fig:rtb-rbst-all}
\end{figure*}

Fig.~\ref{fig:rtb-rbst-all} shows how \texttt{e5b\_E4}, \texttt{e5b\_E2}, \texttt{mpnet\_E4}, \texttt{mpnet\_E2}, \texttt{MiniLM\_E4}, \texttt{MiniLM\_E2}, and the \texttt{OpenAItext} variants adapt to the unseen ARC benchmark.
Fig.~\ref{fig:rtb-rbst-all-h} collects the regret curves of all models implemented by \texttt{Excel\_mask} and \texttt{Excel\_perf\_cost} weighting mechanisms.
Note that Fig.~\ref{fig:rtb-rbst-all-a} is identical to Fig.~\ref{fig:rbst-main-b}, and Fig.~\ref{fig:rtb-rbst-all-g} is identical to Fig.~\ref{fig:rbst-main-a}.
They are included here for completeness.

% \mycmt{Figure comes from \url{https://chatgpt.com/c/68cbf7d5-e4fc-832d-a9a0-5c5c1f1c193a}.}

\vspace{-1mm}\section{MixInstruct Supplementary Materials}
\label{sec:more_mixinst}

\vspace{-2mm}\subsection{Additional Results}
\label{sec:more_mixinst_results}

\begin{figure*}[h]
  \centering
  % Row 1
  \begin{subfigure}{0.48\textwidth}
    \centering
    \includegraphics[width=\linewidth]{plots/mixinst_seed_m1_set1_figure_0.pdf}
    \caption{All text models with top 8\% ambiguous queries removed.}
    \label{fig:mixinst-all-a}
  \end{subfigure}
  \hfill
  \begin{subfigure}{0.48\textwidth}
    \centering
    \includegraphics[width=\linewidth]{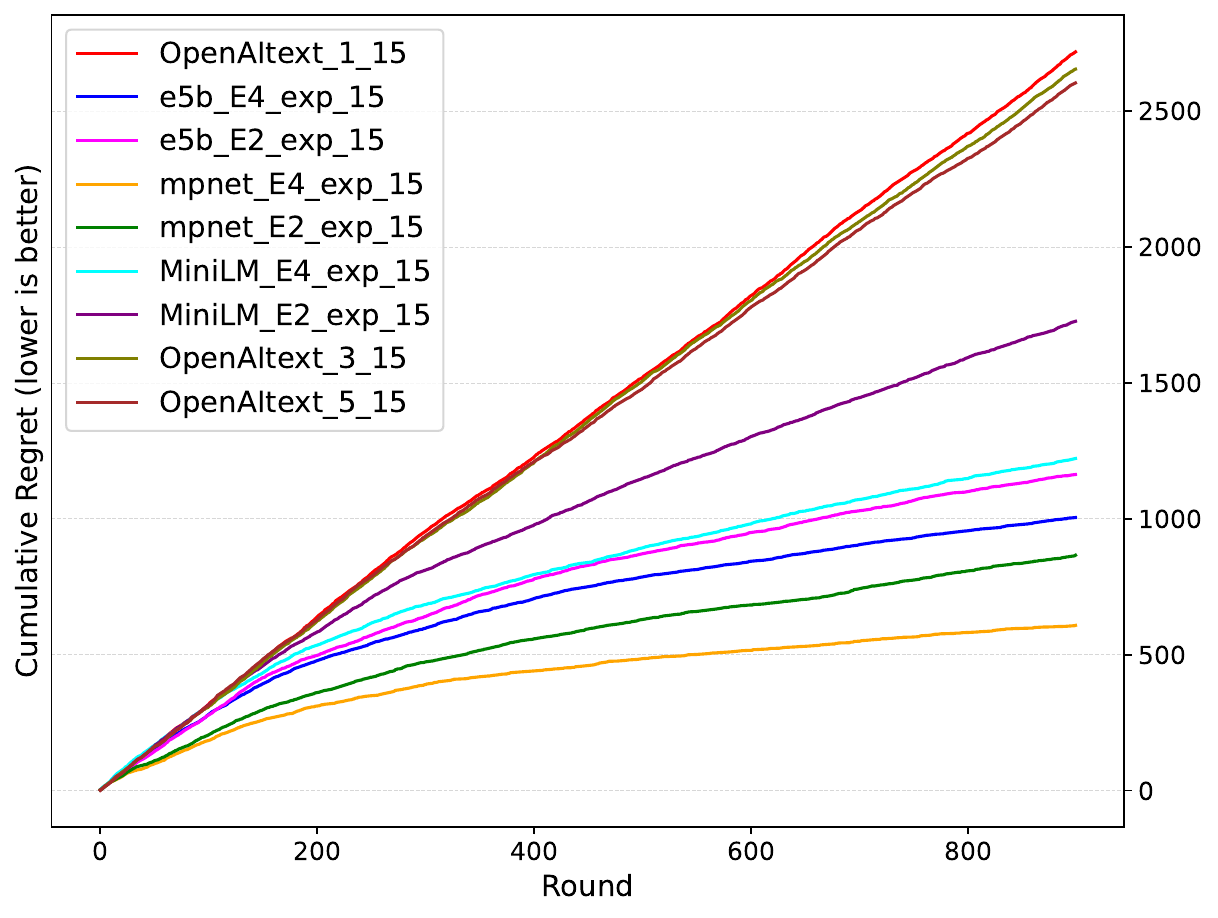}
    \caption{All text models with top 15\% ambiguous queries removed.}
    \label{fig:mixinst-all-b}
  \end{subfigure}

  \vspace{0.3em}

  % Row 2
  \begin{subfigure}{0.48\textwidth}
    \centering
    \includegraphics[width=\linewidth]{plots/mixinst_seed_m1_set2_figure_0.pdf}
    \caption{Ambiguity removal: 8\% versus 15\%.}
    \label{fig:mixinst-all-c}
  \end{subfigure}
  % \hfill
  % \begin{subfigure}{0.48\textwidth}
  %   \centering
  %   \includegraphics[width=\linewidth]{plots/routerb_rbst1_m42_set1_figure_6.pdf}
  %   \caption{H}
  %   \label{fig:mixinst-all-d}
  % \end{subfigure}

  \caption{Cumulative regret curves for MixInstruct.}
  \label{fig:mixinst-all}
\end{figure*}

Figure~\ref{fig:mixinst-all-a}, which is identical to Figure~\ref{fig:mixinst-main-8}, compares all text models with the top 8\% most ambiguous queries removed. 
Figure~\ref{fig:mixinst-all-b} presents results with the top 15\% of ambiguous queries removed. 
Figure~\ref{fig:mixinst-all-c}, identical to Figure~\ref{fig:mixinst-main-compare}, compares the effects of removing different proportions of ambiguous queries.

% \begin{figure}[t]
%   \centering
%   \includegraphics[width=0.5\linewidth]{plots/mixinst_seed_m1_set1_figure_1.pdf}
%   \caption{Cumulative regret curves for MixInstruct.}
%   \label{fig:mixinst-15}
% \end{figure}

% \vspace{-1mm}\section{Prompts for OpenAI's \texttt{text-embedding-3-large} Model}
\vspace{-1mm}\section{Prompts for Naive Implementations in Section~\ref{sec:failure_cases}}
\label{sec:prompts}

The prompt in Listing~\ref{code_snippet:prompt_mmlu} is used to generate model embeddings in \S~\ref{sec:failure_cases}.
The prompt in Listing~\ref{code_snippet:prompt_routerbench} is used to generate model embeddings in \S~\ref{sec:experiments}.

\begin{listing}[h]
\caption{The Python code block including the prompt used in MMLU.}
\label{code_snippet:prompt_mmlu}
\begin{minted}[fontsize=\footnotesize, linenos, breaklines]{python}
prompt = (
    f"This model is very good at solving questions regarding {category}."
    f"Example questions it excels at: "
    f"1. {example_questions[0]}"
    f"2. {example_questions[1]}."
)
\end{minted}
\end{listing}

\begin{listing}[h]
\caption{The Python code block including the prompts used in RouterBench and MixInstruct.}
\label{code_snippet:prompt_routerbench}
\begin{minted}[fontsize=\footnotesize, linenos, breaklines]{python}
avg_perf = np.mean(aggregated_data[model_benchmark]["Perf"])
avg_cost = np.mean(aggregated_data[model_benchmark]["Cost"])
cost_efficiency = 1 / avg_cost if avg_cost > 0 else float("inf")

qs = example_qs[:return_id+1]

if len(qs) > 1:
    questions = ", ".join(qs[:-1]) + f", and {qs[-1]}"
else:
    questions = qs[0]

prompt = (
    f"This is {model_name}, a language model with "
    f"average performance score of {avg_perf:.3f} "
    f"and cost efficiency rating of {cost_efficiency:.3f}."
    f"It has shown particular strength in {model_benchmark} type questions."
    f"Example question(s) it handles: {questions}."
)
\end{minted}
\end{listing}

\section{The Use of Large Language Models}

We used ChatGPT-4o and ChatGPT-5 to assist with the following tasks:
\begin{itemize}
    \item Writing support, including wording suggestions, sentence smoothing, and grammar checking
    \item Table generation
    \item Figure arrangement and layout improvement, including tips for enhancing visualization
    \item Literature review during the initial and drafting stages of the project
\end{itemize}

\end{document}